\documentclass[journal]{IEEEtran}

\usepackage{amsmath,graphicx}
\usepackage{amsfonts}
\usepackage{amsmath,bm}
\usepackage{adjustbox}
\usepackage{algorithmicx}
\usepackage{algpseudocode}
\usepackage[ruled]{algorithm}
\usepackage{graphicx}
\usepackage{caption}
\usepackage{graphicx,times,amsmath} 
\usepackage{caption}
\usepackage{subcaption}
\usepackage[rightcaption]{sidecap}
\usepackage{caption}
\usepackage{multirow}
\usepackage{hhline}
\usepackage{amsmath}
\usepackage{epstopdf}
\usepackage{tabularx}
\usepackage{enumerate}
\usepackage{scrextend}
\usepackage{wrapfig}
\usepackage{caption}
\usepackage{subcaption}
\usepackage{color,soul}
\usepackage[table]{xcolor}

\usepackage{enumitem}

\usepackage{amsfonts}
\usepackage{amsmath}
\usepackage{adjustbox}
\usepackage{algorithmicx}
\usepackage{algpseudocode}
\usepackage[ruled]{algorithm}
\usepackage{graphicx}
\usepackage{caption}
\usepackage{graphicx,times,amsmath} 
\usepackage{makecell}

\usepackage{caption}
\usepackage{subcaption}
\usepackage[rightcaption]{sidecap}
\usepackage{caption}
\usepackage{multirow}
\usepackage{hhline}
\usepackage{amsmath}
\usepackage{epstopdf}
\usepackage{tabularx}
\usepackage{algorithmicx}
\usepackage{algpseudocode}
\usepackage[ruled]{algorithm}

\usepackage{epstopdf}
\usepackage{tabularx}
\usepackage{subcaption}

\usepackage{tabularx,booktabs}
\newcolumntype{C}{>{\centering\arraybackslash}X} 
\usepackage{lipsum}

\makeatletter
\def\old@comma{,}
\catcode`\,=13
\def,{%
  \ifmmode%
    \old@comma\discretionary{}{}{}%
  \else%
    \old@comma%
  \fi%
}
\makeatother

\graphicspath{ {images/} }

\usepackage{alphalph}
\newcommand{\floor}[1]{\lfloor #1 \rfloor}

\title{S-Rocket: Selective Random Convolution Kernels  for Time Series Classification}

\author{Hojjat~Salehinejad, \textit{Member, IEEE}, Yang Wang, Yuanhao Yu, Tang Jin,  and Shahrokh~Valaee, \textit{Fellow, IEEE}
        
\thanks{H. Salehinejad is with the Department of Electrical \& Computer Engineering, University of Toronto, Toronto, Canada. e-mail: hojjat.salehinejad@mail.utoronto.ca;
Y. Wang is with the Noah’s Ark Laboratory, Huawei Technologies Co. Ltd Canada, Toronto, Canada, and Department of Computer Science and Software Engineering, Concordia University, Montreal, Canada. e-mail: wang@concordia.ca;
Y. Yu is with the Noah's Ark Laboratory, Huawei Technologies Co. Ltd Canada, Toronto, Canada. e-mail: yuanhao.yu@huawei.com;
T. Jin is with the Noah's Ark Laboratory, Huawei Technologies Co. Ltd Canada, Toronto, Canada. e-mail: tangjin@huawei.com;
S. Valaee is with the Department of Electrical \& Computer Engineering, University of Toronto, Toronto, Canada. e-mail: valaee@ece.utoronto.ca.

}
}

\begin{document}
\newcommand*{\img}{%
  \includegraphics[
    width=\linewidth,
    height=20pt,
    keepaspectratio=false,
  ]{example-image-a}%
}

\maketitle
%

\begin{abstract}
Random convolution kernel transform (Rocket) is a fast, efficient, and novel approach for time series feature extraction using a large number of independent randomly initialized 1-D convolution kernels of different configurations. The output of the convolution operation on each time series is represented by a \textit{partial positive value} (PPV). A concatenation of PPVs from all kernels is the input feature vector to a Ridge regression classifier. Unlike typical deep learning models, the kernels are not trained and there is no weighted/trainable connection between kernels or concatenated features and the classifier. Since these kernels are generated randomly, a portion of these kernels may not positively contribute in performance of the model. Hence, selection of the most important kernels and pruning the redundant and less important ones is necessary to reduce computational complexity and accelerate inference of Rocket for applications on the edge devices. 
Selection of these kernels is a combinatorial optimization problem. 
In this paper, we propose a scheme for selecting these kernels while maintaining the classification performance. First, the original model is pre-trained at full capacity. Then, a population of binary candidate state vectors is initialized where each element of a vector represents the active/inactive status of a kernel. A population-based optimization algorithm evolves the population in order to find a \textit{best state vector} which minimizes the number of active kernels while maximizing the accuracy of the classifier. This activation function is a linear combination of the total number of active kernels and the classification accuracy of the pre-trained classifier with the active kernels. Finally, the selected kernels in the \textit{best state vector} are utilized to train the Ridge regression classifier with the selected kernels. This approach is evaluated on the standard time series datasets and the results show that on average it can achieve a similar performance to the original models by pruning more than $60\%$ of kernels. In some cases, it can achieve a similar performance using only $1\%$ of the kernels. 

\end{abstract}
\begin{IEEEkeywords}
Convolution kernels, feature selection, population-based optimization, pruning, time series classification.
\end{IEEEkeywords}
\section{Introduction}
\label{sec:intro}
A \textit{regular time series} is generally defined as a sequence of recorded observations through time/index in which the spacing of observation times/indices is constant. In an \textit{irregular time series}, the spacing between observations is not constant. Mostly, \textit{time series} refers to a regular time series in the literature~\cite{fawaz2019deep}. Time series occurs in various real-world applications such as human activity recognition~\cite{yousefi2017survey}, travel mode detection~\cite{soares2019recurrent}, natural language processing~\cite{hirschberg2015advances}, speech recognition~\cite{che2018recurrent}, shopping pattern recognition~\cite{salehinejad2016customer}, electronic health records~\cite{rajkomar2018scalable}, and medical imaging~\cite{salehinejad2021deep}. 

Acknowledging availability of certain massive time series datasets, most recent state-of-the-art methods for time series classification are based on \textit{learning} a large number of parameters such as in recurrent neural networks~\cite{salehinejad2017recent}. Generally, these methods are computationally expensive, challenging to scale, and require significant amount of data and training time~\cite{dempster2020rocket}. Many real-world applications, such as oscillometry signal classification~\cite{XuOA308} and human activity recognition~\cite{yousefi2017survey} applications have access to limited-imbalanced time series data, which further complicates training of \textit{learning} models.

Challenges in training very large learning models for time series have motivated the development of more scalable and much faster time series classification models such as random convolution kernel transform (Rocket)~\cite{dempster2020rocket}, MiniRocket~\cite{dempster2020minirocket}, InceptionTime~\cite{fawaz2020inceptiontime} and Proximity Forest~\cite{lucas2019proximity}. Among these and many other methods, Rocket and MiniRocket have shown significant  performance in classification of time series with noticeably less computational time.

Rocket~\cite{dempster2020rocket} and MiniRocket~\cite{dempster2020minirocket} use random convolution kernels for transforming input time series into a set of features to train a linear classifier, without training the kernels. Both methods have shown fast and accurate time series classification on standard datasets and for different applications in a fraction of the training time of existing methods, such as UCR archive~\cite{dau2019ucr}, inter-burst detection in electroencephalogram (EEG) signals~\cite{lundy2021random}, driver’s distraction detection using EEG signals~\cite{tan2020detecting}, functional near infrared spectroscopy signals classification~\cite{andreu2021single}, and human activity recognition~\cite{salehinejad2022litehar}.

Similar to many machine learning models, Rocket uses a very large number of parameters (not trained), which can be prone to overparameterization if the number of parameters in the model exceeds the size of the training samples~\cite{oymak2020toward}. It is unclear which kernels are more effective in achieving a high classification performance. By identifying these kernels and pruning the redundant and inefficient ones, it is possible to reduce computational complexity of the model for faster inference, particularly on devices with limited resources such as edge devices. 

In general, there are three approaches in solving the overparameterization problem which are structural efficiency, quantization, and pruning~\cite{salehinejad2021energy}. Searching for the best subset of convolution kernels (and hence the corresponding features) is an NP-hard combinatorial optimization problem~\cite{salehinejad2021energy},~\cite{salehinejad2021edropout}. Markov chain Monte Carlo (MCMC)~\cite{dabiri2020replica} and simulated annealing (SA) are two popular methods for solving NP-hard combinatorial problems. We have proposed a binary differential evolution (BDE)~\cite{price2013differential} for pruning deep neural networks~\cite{salehinejad2021edropout}, ~\cite{salehinejad2021pruning}, which can search the optimization landscape in parallel and handle multiple optimization constraints.

Most of the pruning and neural architecture search (NAS) methods are focused on pruning trainable convolution kernels. Unlike typical deep neural networks, the random convolution kernel transform approach in Rocket uses 1-D kernels to generate features from time series without training the kernels. Hence, this structure is different from a multi-layer (deep) feature extraction approach in deep learning. 
In this paper, we propose S-Rocket for reducing the computational complexity of Rocket by selecting the efficient convolution kernels~\cite{dempster2020rocket}.
To best of our knowledge, this is the first attempt in the literature to prune redundant and less efficient kernels for the Rocket models in classification of time series.

S-Rocket has three main steps which are pre-training, optimization, and post-training. First, Rocket is fully trained using the training data. Unlike most typical pruning methods which use a greedy search or statistical evaluation approach, S-Rocket uses the trained classifier at full capacity as the classification accuracy objective function. In the optimization step, similar to our proposed BDE approach in~\cite{salehinejad2020edropout}, a pool of candidate state vectors is used where the binary state of each element in a vector is a representation of an active/inactive kernel. A linear combination of the accuracy of the classifier (trained in step 1) for sparse input features and the number of active kernels per state vector is the objective function to minimize. Finally, the classifier is retrained from scratch with the masked input features by the \textit{best state vector} found in the optimization step\footnote{Our codes are available at: \textit{https://github.com/salehinejad/srocket}}.

\section{Background}

\subsection{Overparameterization}

There are three main non-mutually exclusive approaches to address the overparameterization problem in neural networks which can also be generalized for other machine learning models. These approaches are \textit{structural efficiency}, \textit{quantization}, and \textit{pruning}. The \textit{structural efficiency} approach can be divided into five categories which are \textit{knowledge distillation}, \textit{special matrix structures}, \textit{manually designed architectures}, \textit{neural architecture search}, and \textit{weight sharing}~\cite{salehinejad2021energy}. Some of these approaches are described below. 

\subsubsection{Quantization} The feature values and weights in a model are generally stored as 32-bit floating-point. Storage and computation of these values require dedicated resources and can result in more energy consumption and slower training and inference, particularly for models with a very large number of floating-point operations on resource-limited devices. Quantization can reduce the number of bits used for the representation of the weights and the feature values~\cite{roth2020resource}.

\subsubsection{Manually Designed Architectures} One of the most common approaches for increasing the efficiency of learning models is redesigning different building blocks and architecture of the model. For instance, the global average pooling reduces the spatial dimension of each channel into a single feature by averaging over all values within a channel~\cite{lin2013network}. 
Another example is SqueezeNet~\cite{iandola2016squeezenet}, a redesign of the AlexNet~\cite{krizhevsky2012imagenet}, which reduces the number of channels by implementing $1\times 1$ convolutions.

\subsubsection{Neural Architecture Search}
This approach is broadly used to design models using optimization and search approaches. The optimization problems are generally designed in a discrete space of
possible architectures (states) with an objective (or multi-objective) function~\cite{roth2020resource}. 
Evaluation of the objective function for all possible states is a combinatorial and NP-hard problem~\cite{salehinejad2021energy}.

\subsubsection{Pruning}
\textit{Pruning} refers to permanently removing a subset of a model's parameters. This approach can reduce overfitting (particularly in limited data~\cite{lecun1989optimal}), tackle the overparameterization problem~\cite{hayou2020pruning}, \cite{blalock2020state}, and increase resource efficiency of neural networks. Most pruning methods are applied after fully training a model, followed by retraining the pruned model~\cite{roth2020resource}.
From an architectural perspective, pruning methods are divided into \textit{unstructured} and \textit{structured} approaches.  \textit{Unstructured} pruning does not follow a specific geometry and happens at channel, filter, and intra-filter levels and the geometrically sparse weights are difficult to implement in practice. \textit{Structured} pruning typically follows a geometric structure and removes a subset of weights such as the entire filter~\cite{anwar2017structured} where compared with the \textit{unstructured} approach has very little computational cost overhead~\cite{salehinejad2021energy}.

Pruning generally targets two objectives which are reducing the number of parameters and  increasing the classification performance. Hence, a major challenge is selection of a subset of the model's parameters without dropping the classification performance, which creates a dilemma between performance and model's size. Generally, pruning can reduce the size of a model but may not improve the
efficiency in terms of training or inference time~\cite{cheng2017survey}.

Performance evaluation of neural networks in pruning approaches is generally performed using the Softmax function (which is a form of \textit{Gibbs distribution}). As it is noted in the \textit{Knowledge Distillation} work by Hinton et. al.~\cite{hinton2015distilling}, the temperature factor in the Gibbs distribution can provide a control on softness of the probability distribution over target classes. In addition, computing the partition function is a computing bottleneck, particularly for applications with very large number of target classes (a.k.a. extreme classification problems~\cite{bamler2020extreme})~\cite{barber2016dealing}. The larger the number of classes, the lower the precision of probability values over the target classes, which may encourage learning noise and the gradients vanishing~\cite{bamler2020extreme}, \cite{luo2019mathcal}.

Layer-wise pruning with manual setup of a sensitivity parameter per layer, is a common approach which dictates revisiting and adjusting these parameters during fine-tuning~\cite{cheng2017survey}. One of the main reasons for a layer-wise pruning approach is the lower computational complexity, compared with a network-wise approach.
The common trend to evaluate the importance of parameters is to treat all layers in the network similarly, particularly for threshold-based methods. Pruning all layers uniformly tends to perform worse than \textit{intelligently} allocating parameters to different
layers~\cite{blalock2020state}. The $l_{1}$ and $l_{2}$ norms of the weights are typical methods to detect importance of parameters which requires more iterations to converge than general methods~\cite{cheng2017survey}. \textit{Mutual information} has been used as a metric to measure the strength of the relationship between filters of
adjacent layers, across every pair of layers~\cite{ganesh2020mint}.



\subsection{Evolutionary Pruning}
 Pruning modern learning models is an NP-hard combinatorial optimization problem. As an example, AlexNet~\cite{krizhevsky2012imagenet} has $96$ filters in the first convolutional layer which corresponds to $2^{96}$ pruning possibilities. Deterministic approaches are not able to find a guaranteed solution to this optimization problem. In addition, these methods cannot handle different constraints and pruning criteria in the objective function. Population-based global optimization methods~\cite{onwubolu2009differential} can handle a wide variety of optimization constraints (even at multi-objective~\cite{robivc2005differential} and many-objective~\cite{bandyopadhyay2014algorithm} scales) and can provide at least a feasible solution. 
 
Genetic algorithm (GA) is one of the earliest methods for discovering a combination of connections for enhancing training of multi-layer perceptron models (MLPs)~\cite{whitley1990genetic}. GA, particularly with a small population size, has showed a better performance than the Bayesian models~\cite{cantu2003pruning}. In a multi-objective GA approach~\cite{yang2019multi}, the objective function is to minimize a linear combination of the weighted average of network loss, computational complexity, and sparsity, where each parameter is controlled by a coefficient.

Differential evolution (DE) is another popular population-based method for pruning weights in deep learning models~\cite{salehinejad2020edropout}, \cite{wu2021differential}. We have previously proposed energy-based objective functions for dropout and pruning of deep neural networks using BDE. 
Ising energy objective functions are proposed in~\cite{salehinejad2019ising} and \cite{salehinejad2019ising2}, which represent the saturation and activation level of neurons in an MLP network.
The \textit{EDropout}~\cite{salehinejad2020edropout} method uses energy-based models (EBMs)~\cite{lecun2006tutorial} as a measure
of compatibility, which represents the dependencies of a subset
of the network variables as a scalar energy, based on the
definition of EBMs. In this scheme, a BDE algorithm evolves a set of binary state vectors, to minimize the energy-based objective function. DE is also used in~\cite{wu2021differential} to solve an optimization problem corresponding to sparsity and network accuracy. 

\subsection{Random Convolution Kernel Transform}
It is shown that utilizing the combination of random convolutional filters with rectification, pooling, and local normalization for feature extraction can have a similar performance to learned features in small networks and datasets~\cite{jarrett2009best}. This approach was also implemented in~\cite{yosinski2014transferable} for different number of layers on larger networks and larger datasets, in a dissimilar setup to~\cite{jarrett2009best}. It is shown that the classification performance drops in the first two layers and it gets to near random in layer three and subsequent ones~\cite{yosinski2014transferable}.

\begin{figure}[!t]
\centering
\captionsetup{font=footnotesize}
\begin{subfigure}[t]{0.49\textwidth}
\captionsetup{font=footnotesize}
\centering
\includegraphics[width=1\textwidth]{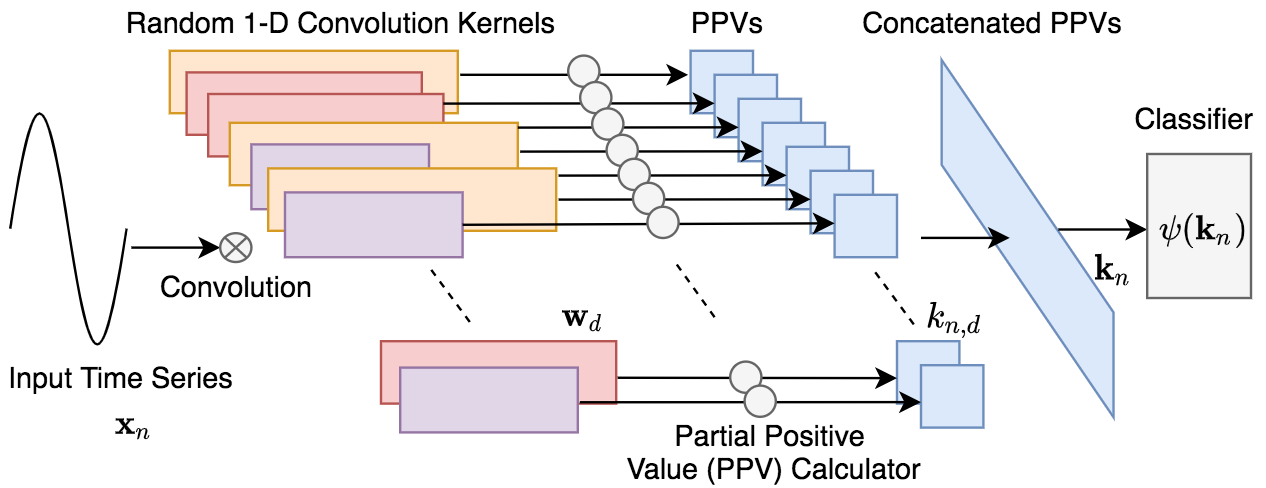}  
\caption{Random convolution kernels for feature extraction and classification.}
\label{fig:rocket_model}
\end{subfigure}%

\begin{subfigure}[t]{0.4\textwidth}
\captionsetup{font=footnotesize}
\centering
\includegraphics[width=1\textwidth]{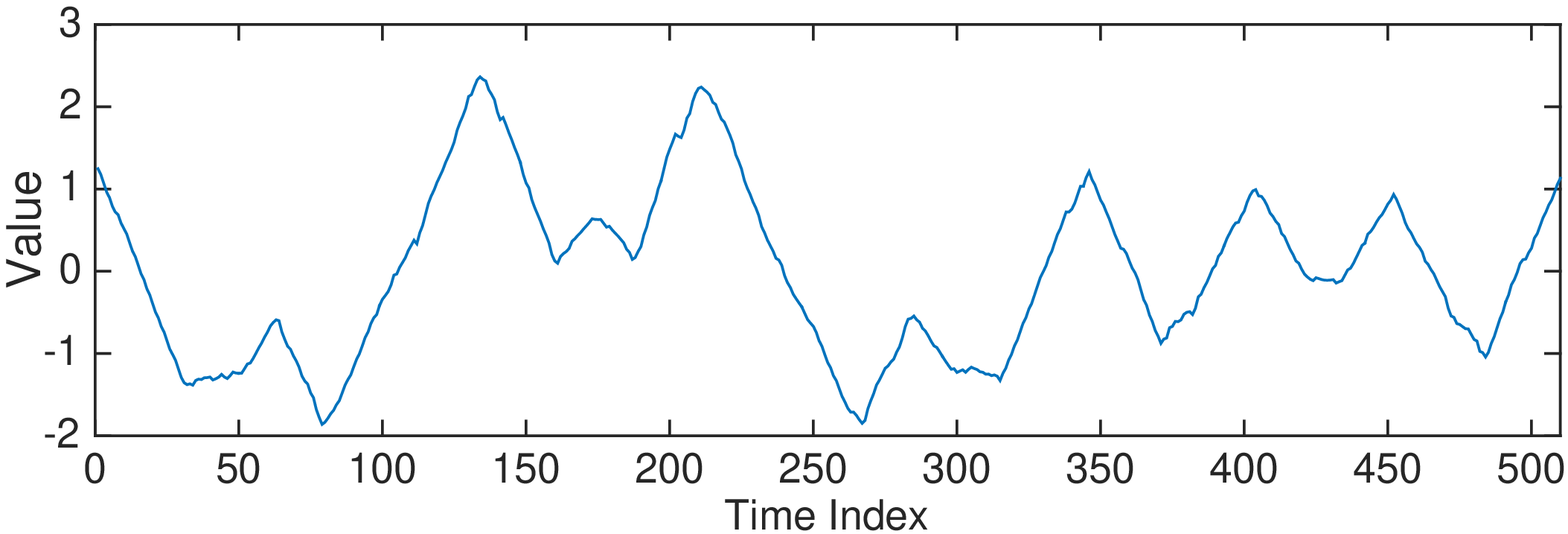}  
\caption{A time series sample $\mathbf{x}_{n}$ from the UCR~\cite{dau2019ucr} dataset.}
\label{fig:sample_time}
\end{subfigure}%

\begin{subfigure}[t]{0.4\textwidth}
\captionsetup{font=footnotesize}
\centering
\includegraphics[width=1\textwidth]{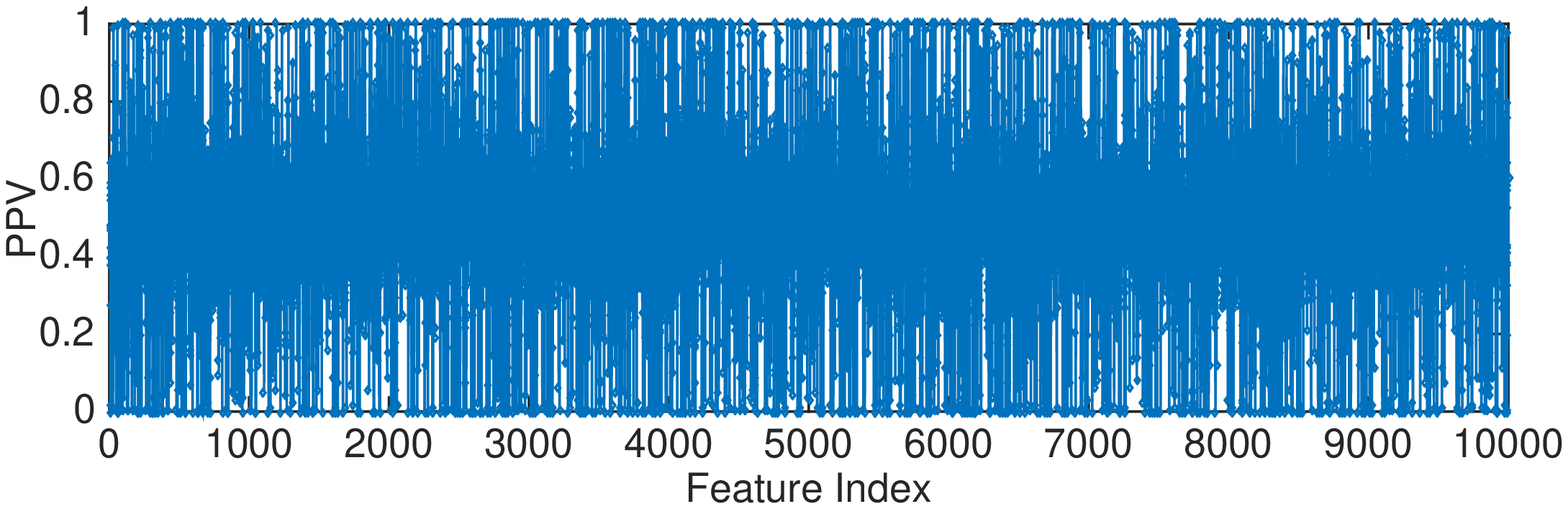}
\caption{The PPV features $\mathbf{k}_{n}$ of the time series in (b).}
\label{fig:sample_ppv}
\end{subfigure}%
\caption{A visualization of the random convolution kernels transform for features extraction and classification. A time series sample and corresponding PPV features from $D=10,000$ random convolution kernels.} 
\label{fig:sample_ppv_time}
\end{figure}

Shallow random networks, initialized as a bank of arbitrary randomized nonlinearities, called Convolutional Kitchen Sinks (CKS)~\cite{rahimi2008weighted}, has achieved good classification results, particularly for transcription factor binding site prediction for DNA sequences using only a one layer random convolutional neural network and a linear classifier~\cite{morrow2017convolutional}. All the parameters of this setup are independent and identically distributed and randomly selected from a Gaussian distribution with a fixed variance.

The conventional training
methods for convolution kernels use gradient-descent searching which is generally a time-consuming task prone to various challenges. In contrast to the learned convolution kernels in typical convolutional neural networks, these kernels are randomly initialized without learning in Rocket models. Rocket leverages different aspects of a convolution kernel with respect to the values of the weights (including the bias term), length, dilation, and padding. 
A bank of randomly generated kernels with respect to these factors, can get astonishing results, particularly with no/few labeled samples~\cite{jarrett2009best}.

Figure~\ref{fig:sample_ppv_time}(a) shows different steps of the Rocket model. It initializes a bank of random convolution kernels (e.g. ${10,000}$ kernels is suggested in~\cite{dempster2020rocket}) where convolution of each kernel with an input time series produces a feature vector. Each feature vector is then represented by the proportion of positive values (PPV) and/or the maximum value~\cite{dempster2020rocket}. Later, the concatenation of PPV values from the kernels is used as the input feature vector to train a Ridge regression classifier. As an example, Figure~\ref{fig:sample_ppv_time}(b) shows a sample time series from the UCR~\cite{dau2019ucr} dataset and the corresponding PPV values are presented in Figure~\ref{fig:sample_ppv_time}(c). This Figure shows that the length of the input feature vector to the classifier is ${10,000}$ which corresponds to the number of 1-D kernels. In fact, regardless of the length of the time series, it is represented by a feature vector with the length equal to the number of kernels. This approach eliminates padding of time series to identical lengths and generalizes the classifier for various-length time series.

One major parameter in initializing kernels is dilation, which works like a sampler that spreads a kernel over an input signal (e.g. in a dilation of three, every third element of the input signal is convolved with the kernel)~\cite{bai2018empirical}. Rocket initializes its weights from a Normal distribution ${\mathcal{N}(0,1)}$ and the bias term from a uniform distribution ${\mathcal{U}(-1,1)}$, 

MiniRocket~\cite{dempster2020minirocket} is similar to Rocket in terms of using random convolution kernels for feature extraction, but with a
small and fixed set of kernels (i.e. $84$ kernels). This approach only uses PPV pooling to represent each feature map while the dilation aspect of Rocket and the inputs are not normalized. The weights in MiniRocket are restricted to specific values in ${\{-1,2\}}$ such that the sum of weights must be zero and the length of the kernels must be restricted to $9$,~\cite{dempster2020minirocket}. The design of MiniRocket follows a \textit{structural efficiency} approach in tackling the overparameterization problem in Rocket, such that it is manually modified for faster and more accurate performance.

\section{S-Rocket Model}
Rocket~\cite{dempster2020rocket} and MiniRocket~\cite{dempster2020minirocket} randomly initialize a large number of convolution kernels for feature extraction. We propose S-Rocket for selecting the most efficient kernels with respect to maintaining the classification performance of the original model while reducing the computational complexity of the original model. S-Rocket is discussed for Rocket and a similar procedure is also applicable to MiniRocket without loss of generality. 

\begin{figure}[!t]
\centering
\captionsetup{font=footnotesize}
\includegraphics[width=0.49\textwidth]{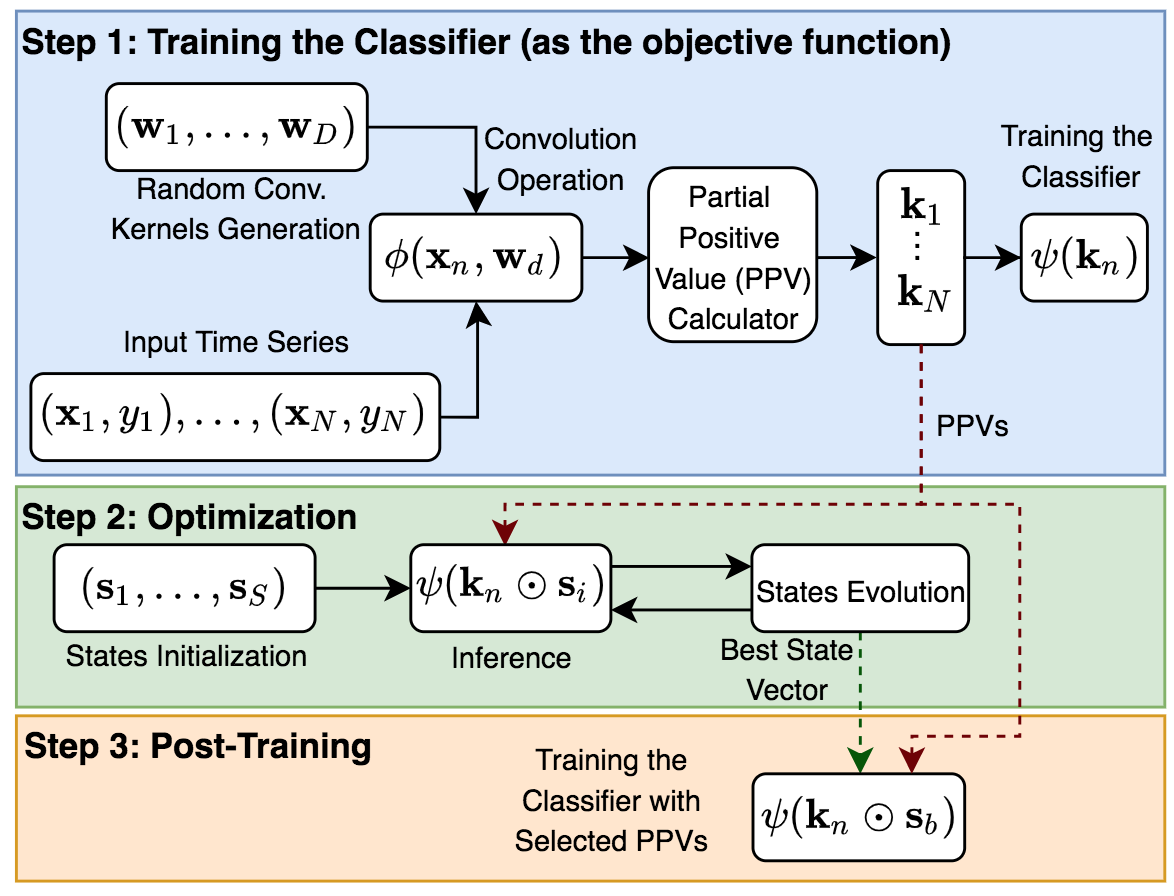}
\caption{Diagram of the pre-training, optimization, and post-training steps in S-Rocket (without loss of generality for S-MiniRocket). $\odot$ is the element-wise product.}
\vspace{-4mm}
\label{fig:s_rocket_model}
\end{figure}

Let $\{(\mathbf{x}_{1},y_{1}),...,(\mathbf{x}_{N},y_{N})\}$ represent the training data, where $\mathbf{x}_{n}$ is a time series and $y_{n}$ is the corresponding label. S-Rocket has three main steps which are pre-training, optimization, and post-training, as demonstrated in Figure~\ref{fig:s_rocket_model}. 
Algorithm~\ref{alg:srocket} also shows the pseudocode of the S-Rocket.

\subsection{Pre-Training}
\subsubsection{Kernels Initialization}
\label{sec:initialization}

A set of $D$ random convolution kernels $\mathbf{\Theta}=(\mathbf{w}_{1},...,\mathbf{w}_{D})$ is initialized, as recommended in~\cite{dempster2020rocket}, such that for each kernel $\mathbf{w}\in\mathbf{\Theta}$ the following values are set:

\begin{itemize}
\item \textit{Length}: The length of each kernel is randomly selected from $\{7, 9, 11\}$ with equal probability.
\item \textit{Weights}: The weights are randomly sampled from a Normal distribution $\mathcal{N}(0,1)$.
\item \textit{Bias}: The bias value for each kernel is randomly chosen from a uniform distribution $\mathcal{U}(-1,1)$;
\item \textit{Dilation}: Sampled from an exponential scale $\kappa=\floor{2^{a}}$, where $a\sim \mathcal{U}(0, log_{2}\frac{|\mathbf{x}|-1}{|\mathbf{w}|-1})$, $|\mathbf{x}|$ is the cardinality (length) of input signal and $|\mathbf{w}|$ is the length of kernel.
\item \textit{Padding}: A binary random decision is made, with equal probability, to apply padding to the input $\mathbf{x}$. With padding, kernels are centered at the first and last indices of $\mathbf{x}$.
\item \textit{Stride}: The stride value is always set to one for all kernels.
\end{itemize}

\subsubsection{Features Extraction}
The two most important
aspects of Rocket in terms of achieving state-of-the-art accuracy
are the use of dilation, sampled on an exponential scale, and the use
of PPV~\cite{dempster2020minirocket}. Hence, we only focus on extracting PPV features in this paper, without loss of generality. 

The convolution operation $\phi(\cdot)$ for a time series $\mathbf{x}$ and convolution kernel $\mathbf{w}$ is
\begin{equation}
\begin{split}
        \mathbf{f}&=\phi(\mathbf{x},\mathbf{w})\\
        &=\Big(\sum_{j=1}^{|\mathbf{w}|}x_{i+j\cdot \kappa}\cdot w_{j}\;\Big|\;i=1,...,|\mathbf{x}|\Big),
\end{split}
\end{equation}
where $\mathbf{f}$ is the extracted feature vector using dilation $\kappa$ and $|\cdot|$ is the cardinality. The PPV value is then computed as
\begin{equation}
    k = \frac{1}{|\mathbf{f}|}\sum_{i=1}^{|\mathbf{f}|}\mathbf{1}[f_{i}>0],
    \label{eq:ppv}
\end{equation}
where $f_i$ is the $i$th element of the vector $\mathbf{f}$, and $\mathbf{1}[f_{i}>0]$ is the indicator function such that ${\mathbf{1}[f_{i}>0]=1}$ if $f_{i}>0$ and $\mathbf{1}[f_{i}>0]=0$ otherwise. Hence, for the input time series $\mathbf{x}_n$, the extracted feature vector using the bank of convolution kernels $\Theta$ is $\mathbf{k}_n=(k_{1},...,k_{D})$.

\subsubsection{Training the Classifier}
Similar to~\cite{dempster2020rocket}, a Ridge regression classifier $\psi(\mathbf{k})$ is trained using the extracted features ${\{(\mathbf{k}_{1},y_1),...,(\mathbf{k}_{N},y_N)\}}$ corresponding to the input time series. 
This classifier does not require extensive hyperparameter setting, except for the regularization parameters which can be set quickly with cross-validation.

\begin{algorithm}[t]
\small
\begin{algorithmic} 
\State // \textbf{Parameters Setup}
\State Set $t=0$ // Optimization counter
\State Set $D=10,000$ // Number of kernels
\State Set $N$ // Number of training samples

\State Initialize $\mathbf{\Theta}=()$ // Empty set of kernels
\State Initialize $\mathbf{K}=()$ // Empty set of PPVs
\State Initialize $\mathbf{S}^{(0)}$ // States initialization according to Section \ref{sec:statesInitialization}

\State // \textbf{Kernels Initialization}
\For{ $d=1 \rightarrow D$} // Kernel counter
\State Initialize the kernel $\mathbf{w}_d$ // Refer to Section~\ref{sec:initialization}
\State Add $\mathbf{w}_d$ to $\mathbf{\Theta}$
\EndFor

\State // \textbf{Feature Extraction}
\For{$n=1\rightarrow N$} // Training sample counter
\State Initialize $\mathbf{k}_n=()$ // Empty set of PPVs per sample
\For{$d=1\rightarrow D$} // Kernel counter
\State Compute $k_{d}$ using (\ref{eq:ppv})
\State Add $k_d$ to $\mathbf{k}_n$
\EndFor
\State Add $\mathbf{k}_n$ to $\mathbf{K}$
\EndFor

\State // \textbf{Pre-Training}
\State Train the classifier $\psi$

\State // \textbf{Optimization}
\State Compute loss of $\mathbf{S}^{(0)}$ as $\mathcal{L}^{(0)}$ using~(\ref{eq:s_rocket_loss})
\For{ $t=1 \rightarrow \mathit{N_{epoch}}$} // epoch counter

\For {$i=1\rightarrow S$} // States counter
\State Generate mutually different $i_{1},i_{2},i_{3}\in \{1,...,S\}$
\For {$d=1\rightarrow D$} // State dimension counter
\State Generate a random number $r_{d}\in[0,1]$
\State Compute mutation vector $v_{i,d}$ using (\ref{eq:mutation})
\EndFor
\State Select candidate state $\tilde{s}^{(t)}_{i}$ using (\ref{eq:crossover})
\EndFor
\State Compute loss of $\tilde{\mathbf{S}}^{(t)}$  using (\ref{eq:s_rocket_loss})
\State Select $\mathbf{S}^{(t)}$ using (\ref{eq:selection})
\State Select the state with the lowest loss from $\mathbf{S}^{(t)}$ as $\mathbf{s}^{(t)}_{b}$
\EndFor

\State // \textbf{Post-Training}
\State Train the classifier $\psi$ using (\ref{eq:maskedKernel}) and $\mathbf{s}^{(t)}_{b}$

\end{algorithmic}
\small
  \caption{S-Rocket}
  \label{alg:srocket}
\end{algorithm}

\subsection{Optimization}
S-Rocket performs structured pruning on the kernels $\mathbf{\Theta}$ for removing redundant and less efficient kernels. 

\subsubsection{Initialization}
\label{sec:statesInitialization}
Let $s_{d}\in\{0,1\}$ represent the activation state of each kernel, which can be extended to the state vector $\mathbf{s}^{1\times D}$ for $D$ kernels. In the initialization step, ${t=0}$, a pool of candidate state vectors $\mathbf{S}^{(t)}\in \mathbb{Z}_{2}^{S\times D}$ is initialized such that $S$ is an even number and denotes the number of candidate state vectors, $s_{i,d}^{(t)}\sim Bernoulli(0.5)$ for ${i\in(1,...,S/2)}$, and $s_{i,d}^{(t)}=1$ for ${i\in(S/2+1,...,S)}$ along all $d \in (1,\ldots, D)$.

\subsubsection{Objective Function}

The optimization objective is to minimize the number of active kernels while maximizing the classification accuracy. Hence, we need to find a \textit{best state vector} $\mathbf{s}_{b}$ by \textit{minimizing} the objective function
\begin{equation}
    \mathcal{L}=\frac{1}{2}(\mathcal{L}_{\mathcal{D}}-\mathcal{L}_{\mathcal{A}}+1),
    \label{eq:s_rocket_loss}
\end{equation}
where $\mathcal{L}$ is the objective function value (OFV) and  $\mathcal{L}_{\mathcal{D}}$ represents the number of active kernels, defined as
\begin{equation}
    \mathcal{L}_{\mathcal{D}} = \sum_{d=1}^{D}s_{d}/D
\end{equation}
and $\mathcal{L}_{\mathcal{A}}$ is the accuracy of the classifier, defined as
\begin{equation}
    \mathcal{L}_{\mathcal{A}} = \frac{1}{N}\sum_{n=1}^{N}\mathbf{1}[y_{n},\underset{c\in\mathcal{C}}{\text{argmax}} (p_{n,1},...,p_{n,C})],
\end{equation}
where $\mathcal{C}=(1,...,C)$ is the set of target classes, $\mathbf{1}[\cdot,\cdot]$ is the identifier function (i.e. $\mathbf{1}[y,c]=1$ if $y=c$ and $\mathbf{1}[y,c]=0$ if $y\neq c$), and 
\begin{equation}
(p_{n,1},...,p_{n,C})=\psi(\mathbf{k}_{n}\odot\mathbf{s}),
\label{eq:maskedKernel}
\end{equation}
where $\odot$ is the element-wise multiplication and $\psi(\cdot)$ is the classifier. It is obvious that since
$0\leq  \mathcal{L}_{A} \leq 1$, and $0 \leq \mathcal{L}_{D}\leq1$, we have
$0\leq  \mathcal{L} \leq 1$.
This objective function is a trade-off between two costs so that ${\cal L}_D $ increases with the number of active kernels and $(1 - {\cal L}_A)$ decreases with a higher classification accuracy.

\begin{table*}[!t]
\captionsetup{font=footnotesize}
\caption{Average classification performance (Avg. Acc.), average objective function value according to~(\ref{eq:s_rocket_loss}) (Avg. OFV), Matthews correlation coefficient (MCC), and average number of kept features (Avg. $D'$), scaled to $[0,1]$ for selecting kernels from \textbf{Rocket}. The reported values are averaged over $10$ independent runs.}
\centering
\begin{adjustbox}{width=1\textwidth}
\centering

\begin{tabular}{c|ccc|cccc|cc|cc|cc}
\hline
\multirow{2}{*}{Dataset} & \multicolumn{3}{c|}{Rocket} & \multicolumn{4}{c|}{S-Rocket} & 
\multicolumn{2}{c|}{Random} &
\multicolumn{2}{c|}{$l_1$-norm~\cite{li2016pruning}} &
\multicolumn{2}{c}{Soft Filter~\cite{he2018soft}} \\ \cline{2-14} 
             & Acc.$\uparrow$ & OFV$\downarrow$ & MCC$\uparrow$
             & Acc.$\uparrow$ & $D' \downarrow$ & OFV$\downarrow$ & MCC$\uparrow$&
             Acc.$\uparrow$ & MCC$\uparrow$& 
             Acc.$\uparrow$ & MCC$\uparrow$& 
             Acc.$\uparrow$ & MCC$\uparrow$
             \\ \hline\hline
Adiac 	&	0.78	&	0.6	&	0.78	&	0.78 (\textbf{=})    	&	0.41	&	0.31	&	0.78	&	0.69	&	0.69	&	0.72	&	0.70	&	0.71	&	0.70	\\ \hline
ArrowHead 	&	0.83	&	0.58	&	0.76	&	0.83 (\textbf{=})    	&	0.41	&	0.29	&	0.75	&	0.79	&	0.70	&	0.82	&	0.74	&	0.82	&	0.75	\\ \hline
Beef      	&	0.82	&	0.59	&	0.80	&	0.82 (\textbf{=})    	&	0.16	&	0.17	&	0.80	&	0.70	&	0.65	&	0.80	&	0.78	&	0.80	&	0.78	\\ \hline
BeetleFly 	&	0.95	&	0.53	&	0.90	&	0.95 (\textbf{=})        	&	0.27	&	0.16	&	0.90	&	0.86	&	0.80	&	0.91	&	0.90	&	0.91	&	0.90	\\ \hline
BirdChicken 	&	0.90	&	0.55	&	0.82	&	0.88 (-0.02)        	&	0.20	&	0.16	&	0.86	&	0.78	&	0.76	&	0.83	&	0.81	&	0.83	&	0.81	\\ \hline
Car 	&	0.90	&	0.55	&	0.86	&	0.87 (-0.02)        	&	0.19	&	0.16	&	0.86	&	0.81	&	0.77	&	0.85	&	0.82	&	0.84	&	0.80	\\ \hline
CBF 	&	1.00	&	0.50	&	0.99	&	1.00 (\textbf{=})        	&	0.19	&	0.10	&	1.00	&	0.85	&	0.78	&	0.97	&	0.94	&	0.99	&	0.97	\\ \hline
CinCECGT 	&	0.81	&	0.60	&	0.79	&	0.80 (-0.01)        	&	0.21	&	0.21	&	0.79	&	0.71	&	0.65	&	0.80	&	0.76	&	0.81	&	0.76	\\ \hline
ChlCon 	&	0.76	&	0.62	&	0.73	&	0.73 (-0.03)        	&	0.31	&	0.29	&	0.72	&	0.62	&	0.54	&	0.73	&	0.71	&	0.73	&	0.71	\\ \hline
Coffee 	&	1.00	&	0.50	&	1.00	&	0.99 (-0.01)        	&	0.58	&	0.29	&	0.99	&	0.95	&	0.91	&	0.98	&	0.96	&	0.98	&	0.96	\\ \hline
Computers 	&	0.77	&	0.61	&	0.59	&	0.77 (\textbf{=})        	&	0.29	&	0.26	&	0.60	&	0.69	&	0.52	&	0.76	&	0.57	&	0.74	&	0.54	\\ \hline
CricketX 	&	0.83	&	0.59	&	0.82	&	0.83 (\textbf{=})        	&	0.72	&	0.45	&	0.82	&	0.81	&	0.79	&	0.82	&	0.81	&	0.82	&	0.81	\\ \hline
CricketY 	&	0.85	&	0.58	&	0.84	&	0.85 (\textbf{=})        	&	0.71	&	0.43	&	0.83	&	0.78	&	0.73	&	0.80	&	0.78	&	0.80	&	0.77	\\ \hline
CricketZ 	&	0.85	&	0.57	&	0.83	&	0.85 (\textbf{=})        	&	0.7	&	0.43	&	0.83	&	0.79	&	0.76	&	0.81	&	0.78	&	0.82	&	0.79	\\ \hline
DiaSizRed 	&	0.95	&	0.52	&	0.94	&	0.95 (\textbf{=})        	&	0.29	&	0.17	&	0.93	&	0.88	&	0.84	&	0.94	&	0.92	&	0.94	&	0.92	\\ \hline
DisPhaOAG 	&	0.75	&	0.62	&	0.71	&	0.75 (\textbf{=})        	&	0.35	&	0.30	&	0.71	&	0.69	&	0.65	&	0.73	&	0.70	&	0.73	&	0.70	\\ \hline
DisPhaOutCor 	&	0.77	&	0.61	&	0.65	&	0.77 (\textbf{=})        	&	0.35	&	0.29	&	0.65	&	0.69	&	0.49	&	0.74	&	0.49	&	0.75	&	0.49	\\ \hline
DoLoDay 	&	0.65	&	0.68	&	0.60	&	0.64 (-0.01)        	&	0.69	&	0.53	&	0.58	&	0.58	&	0.50	&	0.54	&	0.46	&	0.56	&	0.49	\\ \hline
DoLoGam 	&	0.8	&	0.60	&	0.59	&	0.81 (\textbf{+0.01})        	&	0.23	&	0.21	&	0.61	&	0.75	&	0.52	&	0.78	&	0.56	&	0.80	&	0.59	\\ \hline
DoLoWKE 	&	0.97	&	0.51	&	0.94	&	0.94 (-0.03)        	&	0.01	&	0.03	&	0.93	&	0.82	&	0.73	&	0.92	&	0.90	&	0.92	&	0.90	\\ \hline
Earthquakes 	&	0.75	&	0.63	&	0.71	&	0.75 (\textbf{=})        	&	0.01	&	0.13	&	0.71	&	0.69	&	0.65	&	0.75	&	0.71	&	0.75	&	0.71	\\ \hline
ECG200 	&	0.90	&	0.55	&	0.80	&	0.88 (-0.02)        	&	0.18	&	0.15	&	0.80	&	0.81	&	0.68	&	0.89	&	0.80	&	0.89	&	0.80	\\ \hline
ECG5000 	&	0.95	&	0.53	&	0.94	&	0.95 (\textbf{=})        	&	0.29	&	0.17	&	0.94	&	0.73	&	0.61	&	0.89	&	0.86	&	0.89	&	0.87	\\ \hline
ECGFiveDays 	&	1.00	&	0.50	&	1.00	&	1.00 (\textbf{=})        	&	0.21	&	0.1	&	1.00	&	0.89	&	0.86	&	0.98	&	0.96	&	0.98	&	0.96	\\ \hline
EOGHSignal 	&	0.57	&	0.71	&	0.53	&	0.55 (-0.02)        	&	0.56	&	0.51	&	0.53	&	0.49	&	0.48	&	0.55	&	0.51	&	0.55	&	0.51	\\ \hline
EOGVSignal	&	0.44	&	0.78	&	0.41	&	0.44 (\textbf{=})        	&	0.59	&	0.58	&	0.40	&	0.41	&	0.37	&	0.41	&	0.39	&	0.42	&	0.41	\\ \hline
FaceAll 	&	0.79	&	0.60	&	0.78	&	0.80 (\textbf{+0.01})        	&	0.52	&	0.36	&	0.79	&	0.69	&	0.63	&	0.78	&	0.75	&	0.78	&	0.75	\\ \hline
FaceFour 	&	1.00	&	0.50	&	1.00	&	1.00 (\textbf{=})        	&	0.59	&	0.29	&	1.00	&	0.82	&	0.81	&	0.98	&	0.97	&	0.99	&	0.98	\\ \hline
FacesUCR 	&	0.96	&	0.52	&	0.94	&	0.96 (\textbf{=})        	&	0.5	&	0.27	&	0.95	&	0.88	&	0.81	&	0.92	&	0.90	&	0.91	&	0.90	\\ \hline
FiftyWords 	&	0.85	&	0.58	&	0.83	&	0.85 (\textbf{=})        	&	0.78	&	0.46	&	0.83	&	0.76	&	0.69	&	0.66	&	0.51	&	0.65	&	0.51	\\ \hline\hline

\textbf{Average} &    
\textbf{0.84}     &   \textbf{0.58}         &    \textbf{0.80} &
\textbf{0.83 (-0.01)}        &  \textbf{0.39}       &     \textbf{0.26} & \textbf{0.80} &
\textbf{0.75} & \textbf{0.68} &
\textbf{0.80} & \textbf{0.75} &
\textbf{0.80} & \textbf{0.75} \\ \hline


\end{tabular}
\end{adjustbox}
\label{T:srocket_results}
\end{table*}

\begin{table*}[!t]
\captionsetup{font=footnotesize}
\caption{Average classification performance (Avg. Acc.), average objective function value according to~(\ref{eq:s_rocket_loss}) (Avg. OFV), Matthews correlation coefficient (MCC), and average number of kept features (Avg. $D'$), scaled to $[0,1]$ for selecting kernels from \textbf{MiniRocket}. The reported values are averaged over $10$ independent runs.}
\centering
\begin{adjustbox}{width=0.95\textwidth}
\centering

\begin{tabular}{c|ccc|cccc|cc|cc|cc}
\hline
\multirow{2}{*}{Dataset} & \multicolumn{3}{c|}{Rocket} & \multicolumn{4}{c|}{S-Rocket} & 
\multicolumn{2}{c|}{Random} &
\multicolumn{2}{c|}{$l_1$-norm~\cite{li2016pruning}} &
\multicolumn{2}{c}{Soft Filter~\cite{he2018soft}} \\ \cline{2-14} 
             & Acc.$\uparrow$ & OFV$\downarrow$ & MCC$\uparrow$
             & Acc.$\uparrow$ & $D' \downarrow$ & OFV$\downarrow$ & MCC$\uparrow$&
             Acc.$\uparrow$ & MCC$\uparrow$& 
             Acc.$\uparrow$ & MCC$\uparrow$& 
             Acc.$\uparrow$ & MCC$\uparrow$
             \\ \hline\hline
Adiac 	&	0.82	&	0.59	&	0.76	&	0.78(-0.04)	&	0.10	&	0.16	&	0.75	&	0.63	&	0.60	&	0.71	&	0.70	&	0.72	&	0.70	\\ \hline
ArrowHead 	&	0.87	&	0.57	&	0.85	&	0.86(-0.01)	&	0.35	&	0.25	&	0.84	&	0.81	&	0.78	&	0.83	&	0.81	&	0.83	&	0.81	\\ \hline
Beef      	&	0.87	&	0.57	&	0.87	&	0.83(-0.04)	&	0.10	&	0.14	&	0.82	&	0.73	&	0.69	&	0.79	&	0.77	&	0.81	&	0.80	\\ \hline
BeetleFly 	&	0.88	&	0.56	&	0.84	&	0.87(-0.01)	&	0.30	&	0.21	&	0.84	&	0.79	&	0.70	&	0.86	&	0.84	&	0.86	&	0.84	\\ \hline
BirdChicken 	&	0.90	&	0.55	&	0.88	&	0.89(-0.01)	&	0.19	&	0.15	&	0.86	&	0.81	&	0.80	&	0.84	&	0.82	&	0.86	&	0.83	\\ \hline
Car 	&	0.92	&	0.54	&	0.90	&	0.92(\textbf{=})	&	0.49	&	0.28	&	0.90	&	0.78	&	0.74	&	0.90	&	0.87	&	0.91	&	0.87	\\ \hline
CBF 	&	1.00	&	0.50	&	1.00	&	1.00(\textbf{=})	&	0.01	&	0.01	&	1.00	&	0.89	&	0.87	&	0.96	&	0.94	&	0.98	&	0.96	\\ \hline
CinCECGT 	&	0.87	&	0.57	&	0.86	&	0.84(-0.03)	&	0.39	&	0.28	&	0.81	&	0.65	&	0.53	&	0.75	&	0.72	&	0.76	&	0.75	\\ \hline
ChlCon 	&	0.76	&	0.62	&	0.75	&	0.72(-0.04)	&	0.30	&	0.29	&	0.70	&	0.53	&	0.50	&	0.72	&	0.69	&	0.72	&	0.71	\\ \hline
Coffee 	&	1.00	&	0.50	&	1.00	&	1.00(\textbf{=})	&	0.33	&	0.17	&	1.00	&	0.87	&	0.78	&	0.99	&	0.98	&	0.99	&	0.98	\\ \hline
Computers 	&	0.72	&	0.64	&	0.70	&	0.73(\textbf{+0.01})	&	0.34	&	0.31	&	0.72	&	0.64	&	0.63	&	0.71	&	0.68	&	0.69	&	0.64	\\ \hline
CricketX 	&	0.82	&	0.59	&	0.80	&	0.79(-0.03)	&	0.19	&	0.20	&	0.77	&	0.66	&	0.60	&	0.71	&	0.70	&	0.73	&	0.70	\\ \hline
CricketY 	&	0.83	&	0.58	&	0.81	&	0.81(-0.02)	&	0.20	&	0.19	&	0.80	&	0.76	&	0.70	&	0.80	&	0.77	&	0.79	&	0.77	\\ \hline
CricketZ 	&	0.83	&	0.59	&	0.80	&	0.82(-0.01)	&	0.62	&	0.40	&	0.80	&	0.73	&	0.70	&	0.80	&	0.78	&	0.81	&	0.79	\\ \hline
DiaSizRed 	&	0.93	&	0.54	&	0.92	&	0.93(\textbf{=})	&	0.54	&	0.31	&	0.92	&	0.85	&	0.79	&	0.90	&	0.89	&	0.90	&	0.90	\\ \hline
DisPhaOAG 	&	0.75	&	0.63	&	0.69	&	0.75(\textbf{=})	&	0.27	&	0.27	&	0.70	&	0.58	&	0.48	&	0.71	&	0.70	&	0.72	&	0.70	\\ \hline
DisPhaOutCor 	&	0.78	&	0.61	&	0.75	&	0.78(\textbf{=})	&	0.33	&	0.28	&	0.77	&	0.69	&	0.62	&	0.76	&	0.74	&	0.76	&	0.74	\\ \hline
DoLoDay 	&	0.59	&	0.7	&	0.55	&	0.59(\textbf{=})	&	0.73	&	0.57	&	0.55	&	0.34	&	0.29	&	0.51	&	0.47	&	0.53	&	0.48	\\ \hline
DoLoGam 	&	0.84	&	0.58	&	0.83	&	0.84(\textbf{=})	&	0.40	&	0.29	&	0.82	&	0.68	&	0.63	&	0.80	&	0.76	&	0.80	&	0.77	\\ \hline
DoLoWKE 	&	0.98	&	0.51	&	0.97	&	0.97(\textbf{=})	&	0.01	&	0.02	&	0.96	&	0.90	&	0.85	&	0.95	&	0.94	&	0.95	&	0.94	\\ \hline
Earthquakes 	&	0.75	&	0.63	&	0.74	&	0.75(\textbf{=})	&	0.01	&	0.13	&	0.72	&	0.59	&	0.52	&	0.73	&	0.68	&	0.74	&	0.70	\\ \hline
ECG200 	&	0.92	&	0.54	&	0.90	&	0.91(-0.01)	&	0.20	&	0.15	&	0.90	&	0.84	&	0.82	&	0.89	&	0.88	&	0.90	&	0.88	\\ \hline
ECG5000 	&	0.94	&	0.53	&	0.93	&	0.94(\textbf{=})	&	0.39	&	0.22	&	0.93	&	0.81	&	0.79	&	0.93	&	0.89	&	0.93	&	0.89	\\ \hline
ECGFiveDays 	&	1.00	&	0.50	&	1.00	&	1.00(\textbf{=})	&	0.21	&	0.10	&	1.00	&	0.86	&	0.83	&	0.97	&	0.95	&	0.98	&	0.96	\\ \hline
EOGHSignal 	&	0.60	&	0.70	&	0.52	&	0.59(-0.01)	&	0.29	&	0.35	&	0.54	&	0.46	&	0.40	&	0.56	&	0.55	&	0.58	&	0.56	\\ \hline
EOGVSignal	&	0.54	&	0.73	&	0.49	&	0.52(\textbf{=})	&	0.37	&	0.42	&	0.48	&	0.41	&	0.34	&	0.48	&	0.42	&	0.47	&	0.42	\\ \hline
FaceAll 	&	0.81	&	0.59	&	0.80	&	0.81(\textbf{=})	&	0.72	&	0.46	&	0.80	&	0.71	&	0.69	&	0.79	&	0.77	&	0.79	&	0.77	\\ \hline
FaceFour 	&	0.99	&	0.51	&	0.99	&	0.99(\textbf{=})	&	0.67	&	0.34	&	0.99	&	0.84	&	0.80	&	0.96	&	0.94	&	0.96	&	0.95	\\ \hline
FacesUCR 	&	0.96	&	0.52	&	0.95	&	0.96(\textbf{=})	&	0.80	&	0.42	&	0.96	&	0.86	&	0.84	&	0.92	&	0.90	&	0.91	&	0.90	\\ \hline
FiftyWords 	&	0.84	&	0.58	&	0.83	&	0.84(\textbf{=})	&	0.78	&	0.47	&	0.83	&	0.77	&	0.71	&	0.81	&	0.80	&	0.81	&	0.80	\\ \hline\hline

\textbf{Average} &    
\textbf{0.84}     &   \textbf{0.58}         &    \textbf{0.82} &
\textbf{0.83 (-0.01)}        &  \textbf{0.36}       &     \textbf{0.27} & \textbf{0.82} &
\textbf{0.72} & \textbf{0.67} &
\textbf{0.80} & \textbf{0.78} &
\textbf{0.81} & \textbf{0.78} \\ \hline

\end{tabular}
\end{adjustbox}
\label{T:sminirocket_results}
\end{table*}

\subsubsection{Searching for the Best State Vector}
Since $D$ is generally a large number (e.g. $D=10,000$ as recommended in \cite{dempster2020rocket}), $2^D$ possible state vectors exist. Hence, searching for a \textit{best state vector} which corresponds to the minimum objective function value is an NP-hard combinatorial problem.
Motivated by~\cite{salehinejad2021edropout} and \cite{salehinejad2021framework}, we propose a global optimization approach to minimize (\ref{eq:s_rocket_loss}). 

As Algorithm~\ref{alg:srocket} shows, a mutation vector is computed at each optimization epoch $t$ for each candidate state vector ${\mathbf{s}_{i}^{(t-1)}\in\mathbf{S}^{(t-1)}}$ and all $d\in(1,..,D)$ as 
\begin{equation}
v_{i,d}=\begin{cases}
               1-s_{i_{1},d}^{(t-1)}, \:\:\:\:$if$\:\:\:s_{i_{2},d}^{(t-1)}\neq s_{i_{3},d} ^{(t-1)}\;$\&$\; r_{d}<F\\
               s_{i_{1},d}^{(t-1)}, \:\:\:\:\:\:\:\:\:\:\:$ otherwise $
            \end{cases},
\label{eq:mutation}
\end{equation}
where $F$ is the mutation factor~\cite{salehinejad2017micro}, $r_{d}\in[0,1]$ is a random number, and $i_{1},i_{2},i_{3}\in (1,...,S)$ are  different indices in the pool of candidate kernels. Then, the crossover  operation is defined as
\begin{equation}
\tilde{s}^{(t)}_{i,d}=\begin{cases}
               v_{i,d} \:\:\:\:\:\:\:\:\:\:\:\:$if$\:\:\: r'_{d}\in[0,1] \leq C_{r}\\
               s_{i,d}^{(t-1)} \:\:\:\:\:\:\:\:\:\:\:$ otherwise $
            \end{cases},
\label{eq:crossover}
\end{equation}
 where $C_{r}$ is the crossover coefficient~\cite{salehinejad2017micro}. The parameters $C_{r}$ and $F$ control exploration and exploitation of the pool of candidate state vectors on the optimization landscape. The objective function value of each state vector $\tilde{\mathbf{s}}_{i}^{(t)}$ is then compared with its corresponding parent using (\ref{eq:s_rocket_loss}) as
\begin{equation}
\mathbf{s}_{i}^{(t)}=\begin{cases}
               \tilde{\mathbf{s}}_{i}^{(t)} \:\:\:\:\:\:\:\:\:\:$if$\:\:\:    \mathcal{L}(\tilde{\mathbf{s}}_{i}^{(t)})\leq \mathcal{L}(\mathbf{s}_{i}^{(t-1)}) \\
              \mathbf{s}_{i}^{(t-1)} \:\:\:\:\:$ otherwise $
            \end{cases},
\label{eq:selection}
\end{equation}
for all $i\in(1,...,S)$.

In each epoch, the \textit{best state vector} $\mathbf{s}_{b}^{(t)}$ is selected where
\begin{equation}
  b=\underset{b\in(1,...,S)}{\text{argmin}}\big(\mathcal{L}(\mathbf{s}^{(t)}_{1}),...,\mathcal{L}(\mathbf{s}^{(t)}_{S})\big)  
\end{equation}
and its corresponding objective function value is
\begin{equation}
    \mathcal{L}(\mathbf{s}_{b}^{(t)})=\text{min}\big(\mathcal{L}(\mathbf{s}^{(t)}_{1}),...,\mathcal{L}(\mathbf{s}^{(t)}_{S})\big),
\end{equation}
where the number of active kernels (features) is $D'=|\mathbf{s}_{b}^{(t)}|$ and $|\cdot|$ is the cardinality.

\subsubsection{Termination Criteria}
\label{sec:earlystateconvergence}
The optimization process terminates after $N_{epoch}$ epochs or upon \textit{states convergence} where all the states in $\mathbf{S}^{(t)}$  converge to a state $\mathbf{s}_{b}\in\mathbf{S}^{(t)}$. This can be detected by 
\begin{equation} 
\Delta\mathcal{L} =
 \mathcal{L}(\mathbf{s}_{b}^{(t)}) -
\frac{1}{S}\sum\limits_{j=1}^{S}\mathcal{L}(\mathbf{s}_{j}^{(t)}),
\label{eq:stateconvergence}
\end{equation}
where if $\Delta\mathcal{L}=0$ the optimization process terminates~\cite{salehinejad2021edropout}.

\begin{figure*}[!t]
\centering
\captionsetup{font=footnotesize}
\begin{subfigure}[t]{0.4\textwidth}
\captionsetup{font=footnotesize}
\centering
\includegraphics[width=1\textwidth]{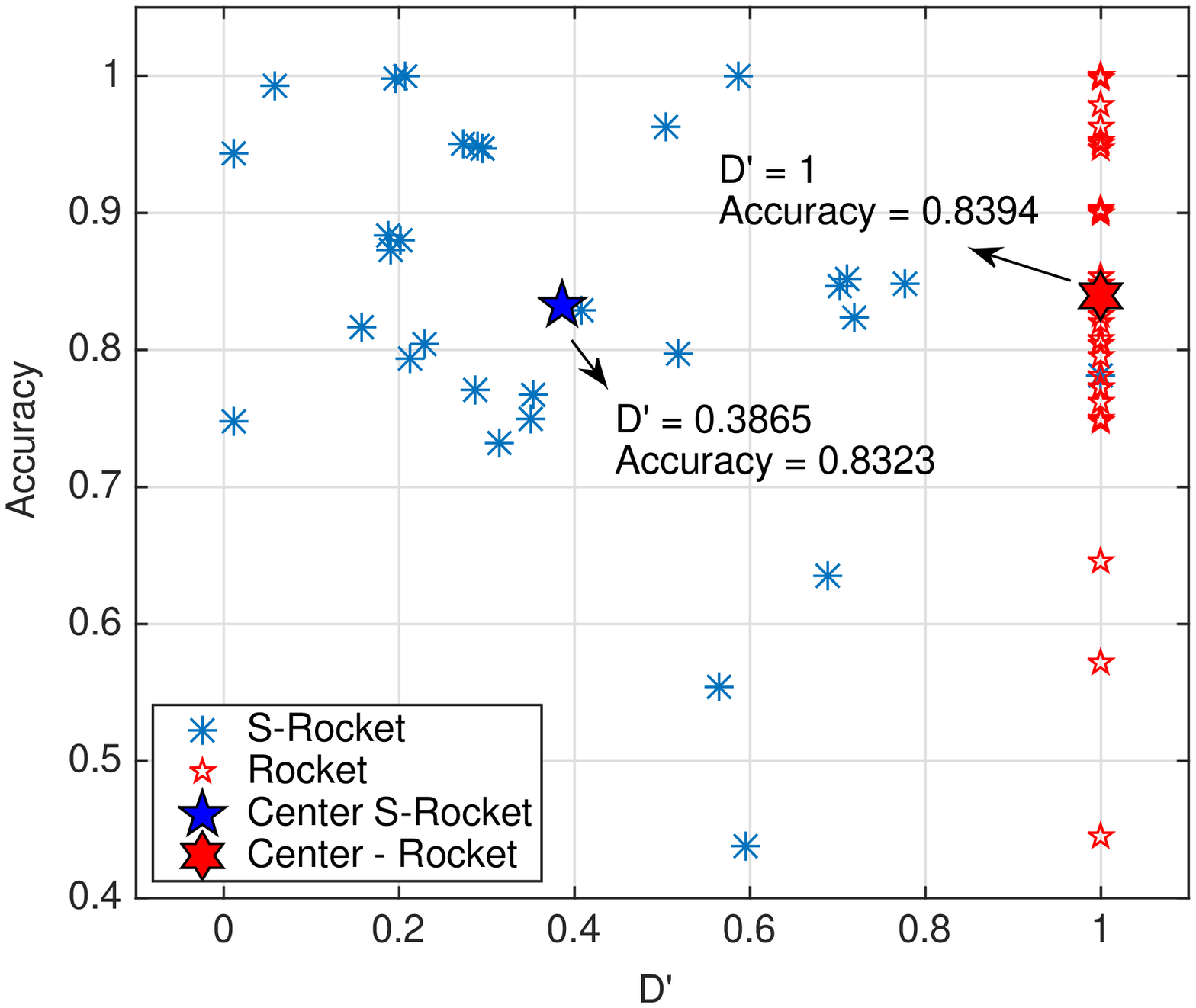}  
\caption{Rocket}
\label{fig:scatterRocket}
\end{subfigure}%
\begin{subfigure}[t]{0.4\textwidth}
\captionsetup{font=footnotesize}
\centering
\includegraphics[width=1\textwidth]{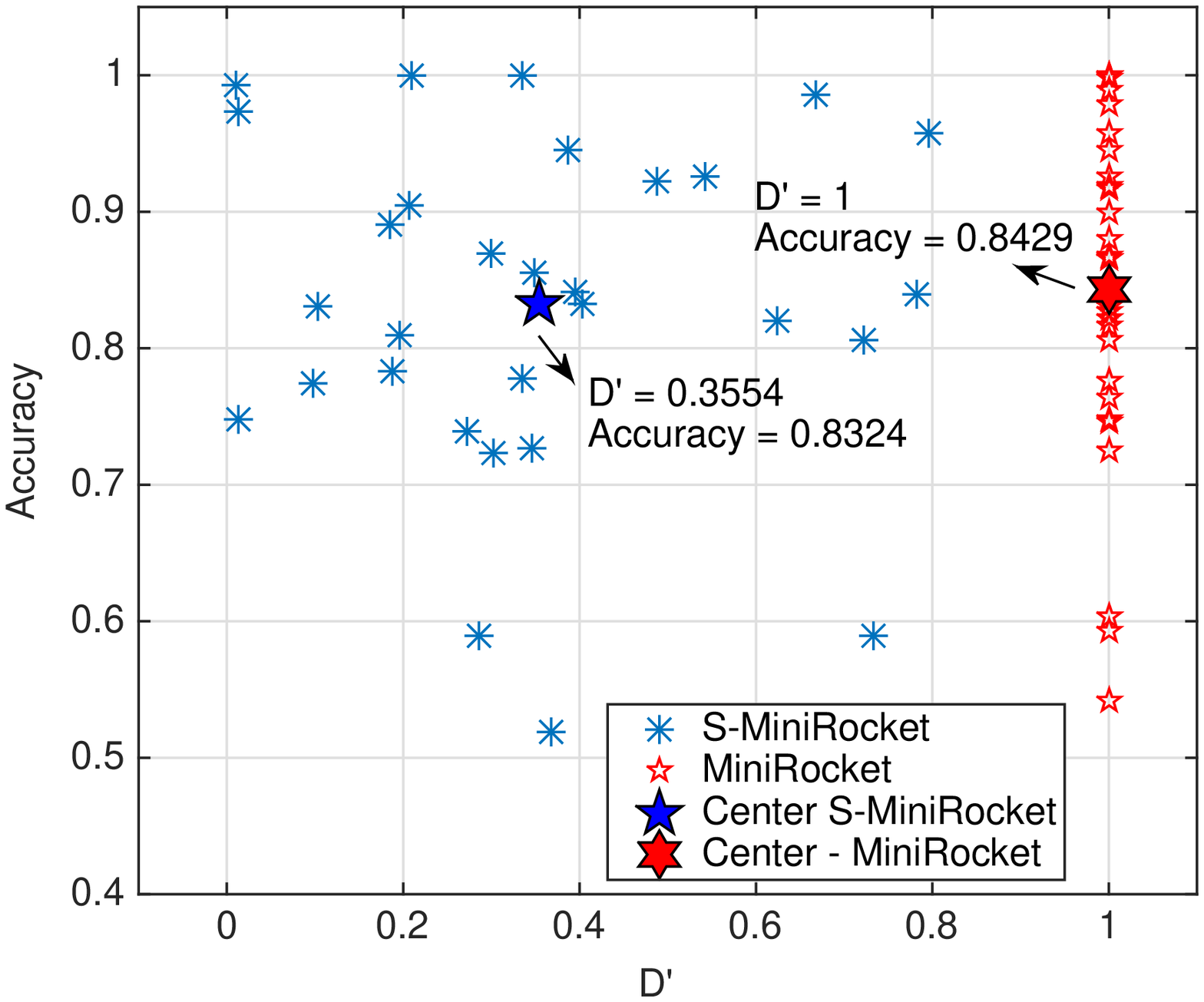}
\caption{MiniRocket}
\label{fig:scatterMiniRocket}
\end{subfigure}%
\caption{Classification accuracy vs. ratio of active convolution kernels ($D'$) for Rocket, MiniRocket, and the corresponding pruned models.} 
\label{fig:scatter_results}
\vspace{-4mm}
\end{figure*}

\begin{figure*}[!t]
\centering
\captionsetup{font=footnotesize}
\begin{subfigure}[t]{0.25\textwidth}
\captionsetup{font=footnotesize}
\centering
\includegraphics[width=1\textwidth]{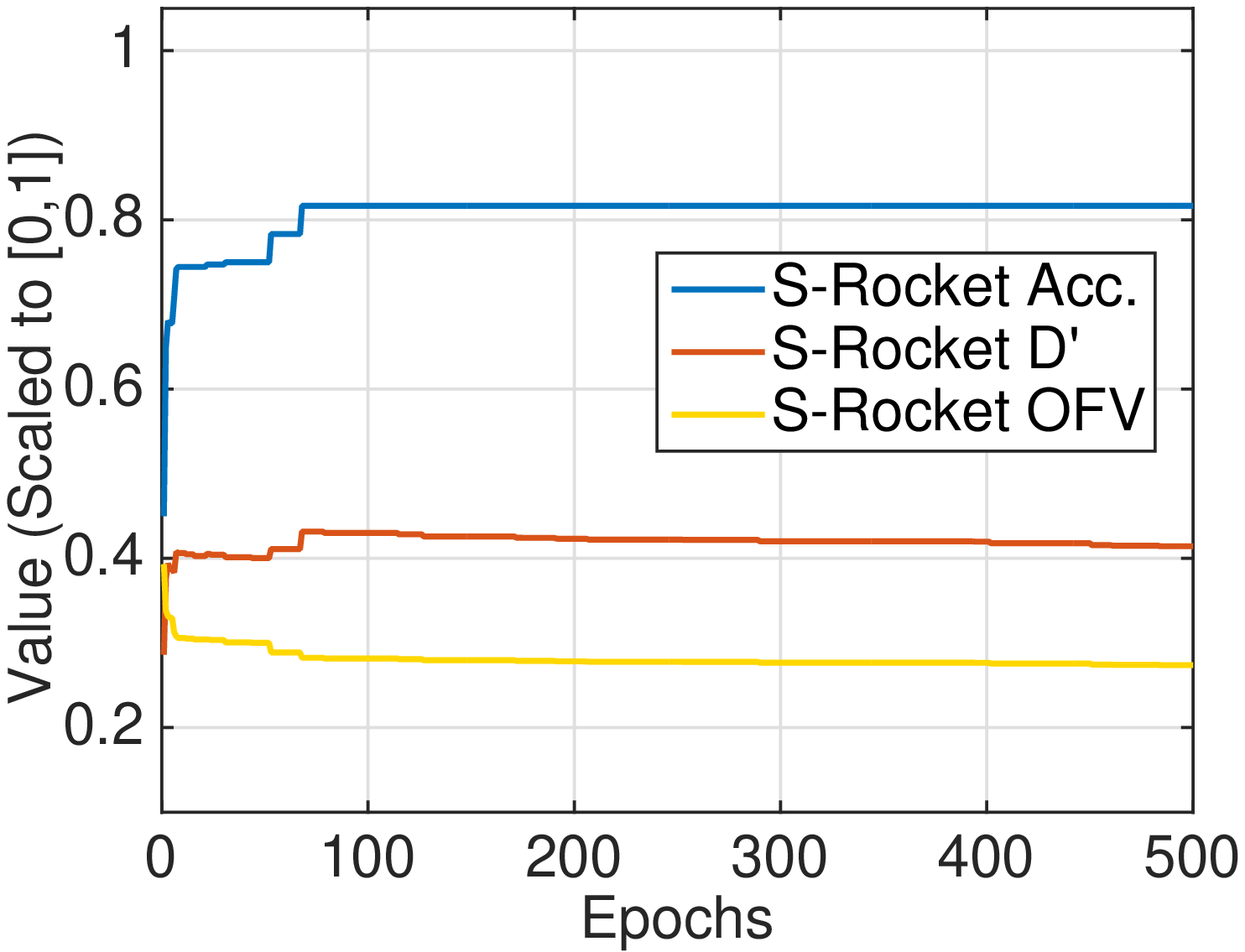} 
\caption{ArrowHead - Rocket}
\label{fig:arrowheadConv}
\end{subfigure}%
\begin{subfigure}[t]{0.25\textwidth}
\captionsetup{font=footnotesize}
\centering
\includegraphics[width=1\textwidth]{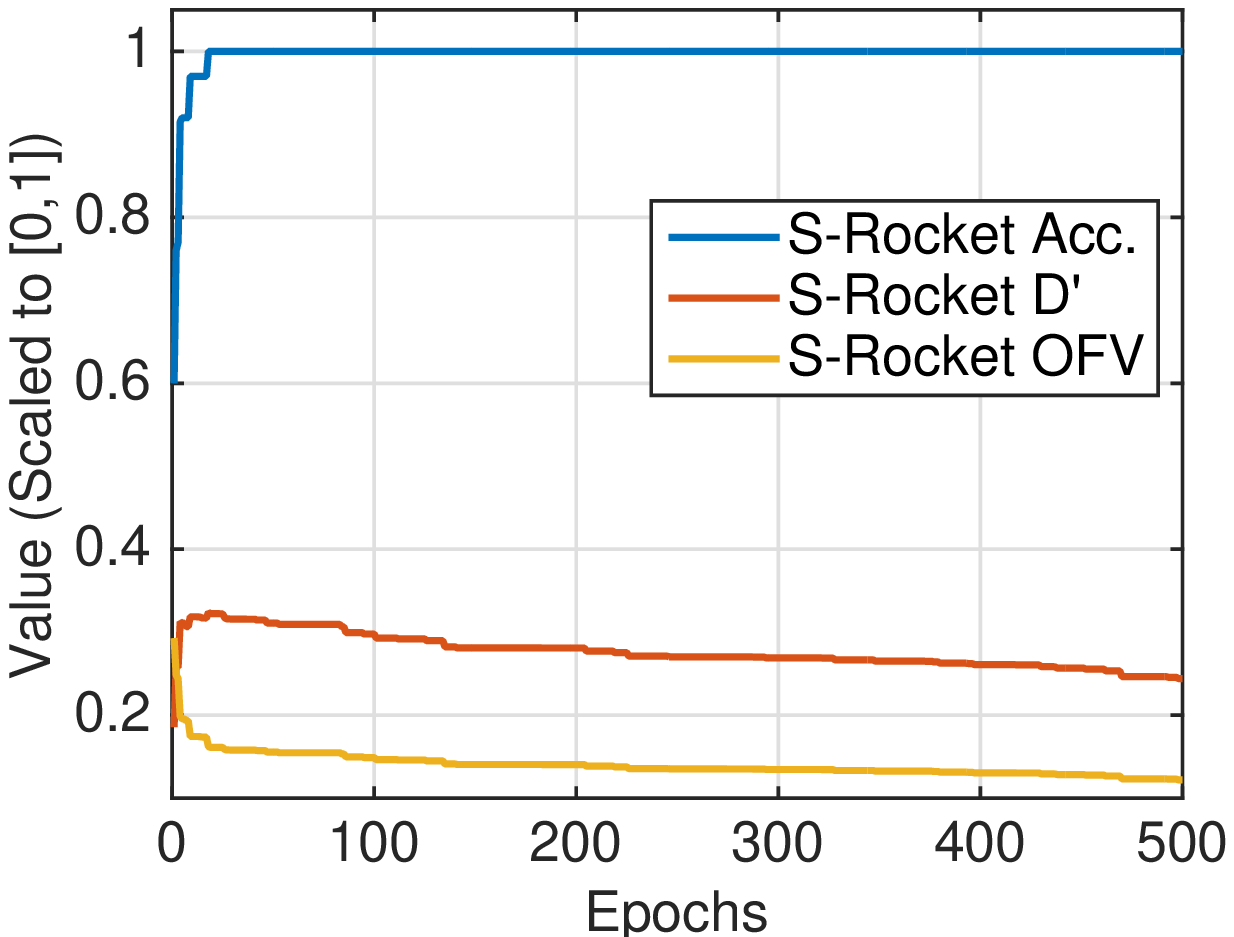}
\caption{BeetleFly - Rocket}
\label{fig:aBeetleFlyConv}
\end{subfigure}%
\begin{subfigure}[t]{0.25\textwidth}
\captionsetup{font=footnotesize}
\centering
\includegraphics[width=1\textwidth]{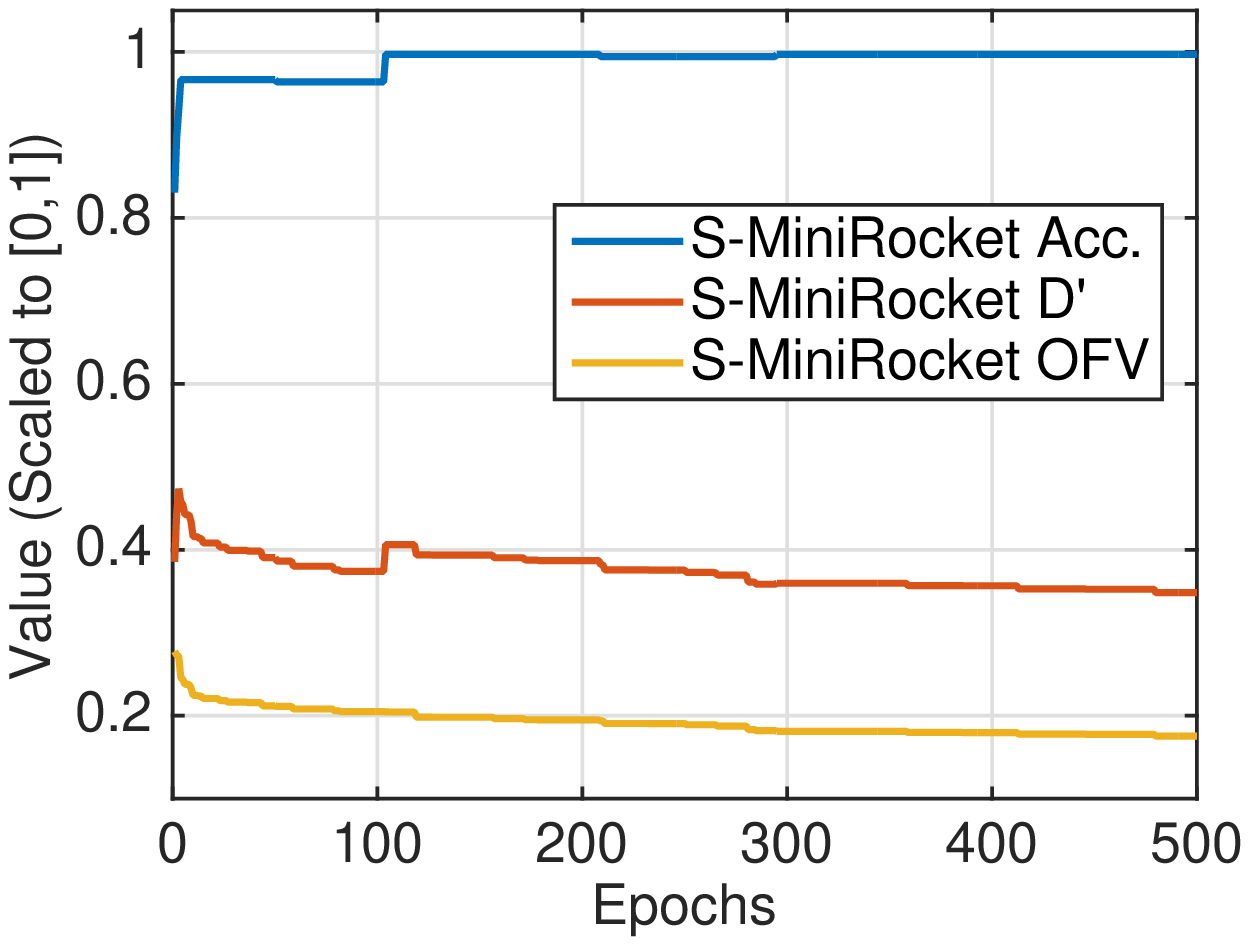}
\caption{ArrowHead - MiniRocket}
\label{fig:arrowheadConvMini}
\end{subfigure}%
\begin{subfigure}[t]{0.25\textwidth}
\captionsetup{font=footnotesize}
\centering
\includegraphics[width=1\textwidth]{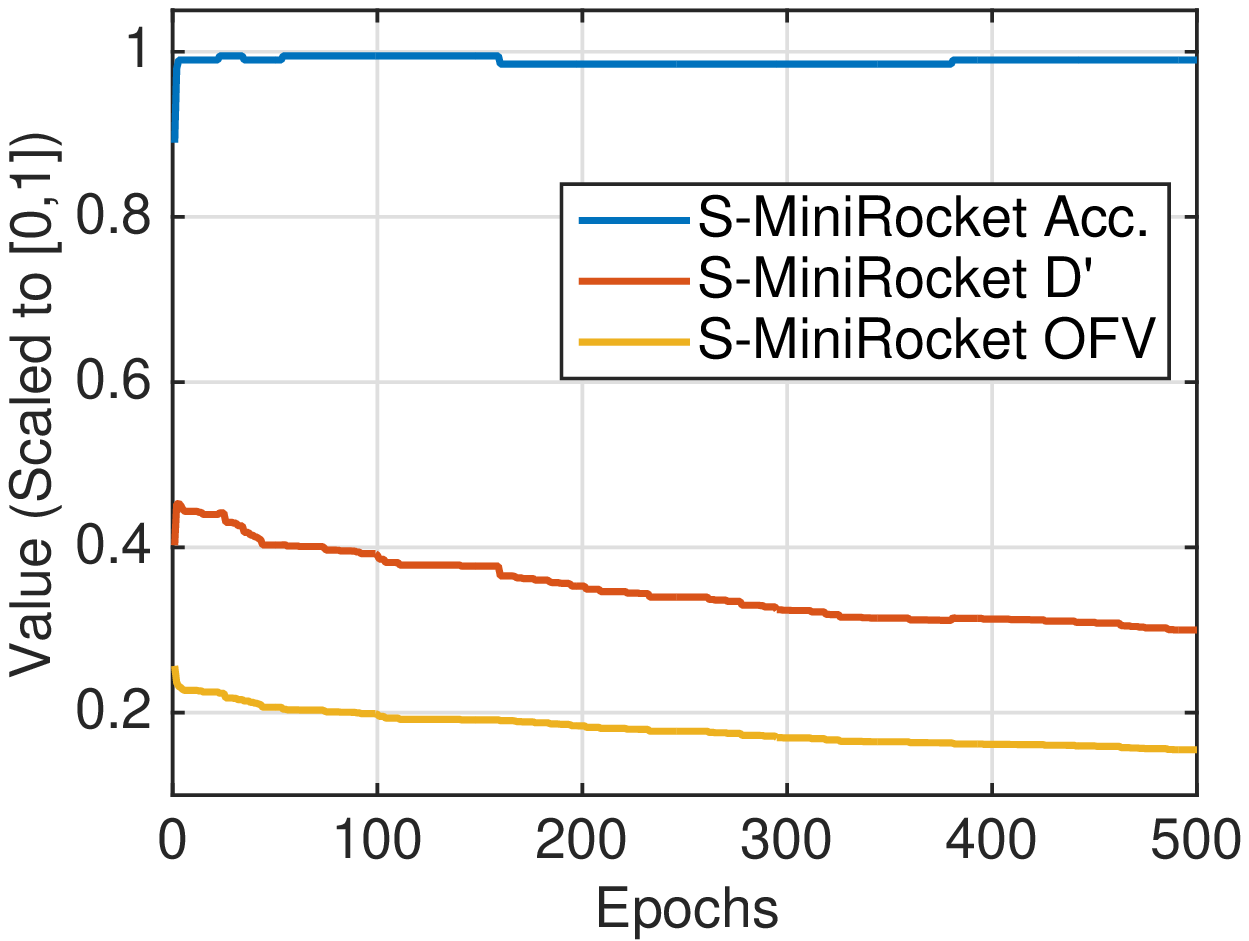}
\caption{BeetleFly - MiniRocket}
\label{fig:aBeetleFlyConvMini}
\end{subfigure}%
\caption{Convergence plots of S-Rocket and S-MiniRocket during $500$ optimization epochs.
OFV: The state vector with lowest objective function value according
to (3) in each epoch; Acc.: The state vector with highest classification accuracy in each epoch; $D'$:
The state vector with lowest number of selected convolution kernels in each epoch. The values are averaged over $10$ independent runs.} 
\label{fig:convergence}
\end{figure*}

\begin{figure}[!t]
\centering
\captionsetup{font=footnotesize}
\begin{subfigure}[t]{0.43\textwidth}
\captionsetup{font=footnotesize}
\centering
\includegraphics[width=1\textwidth]{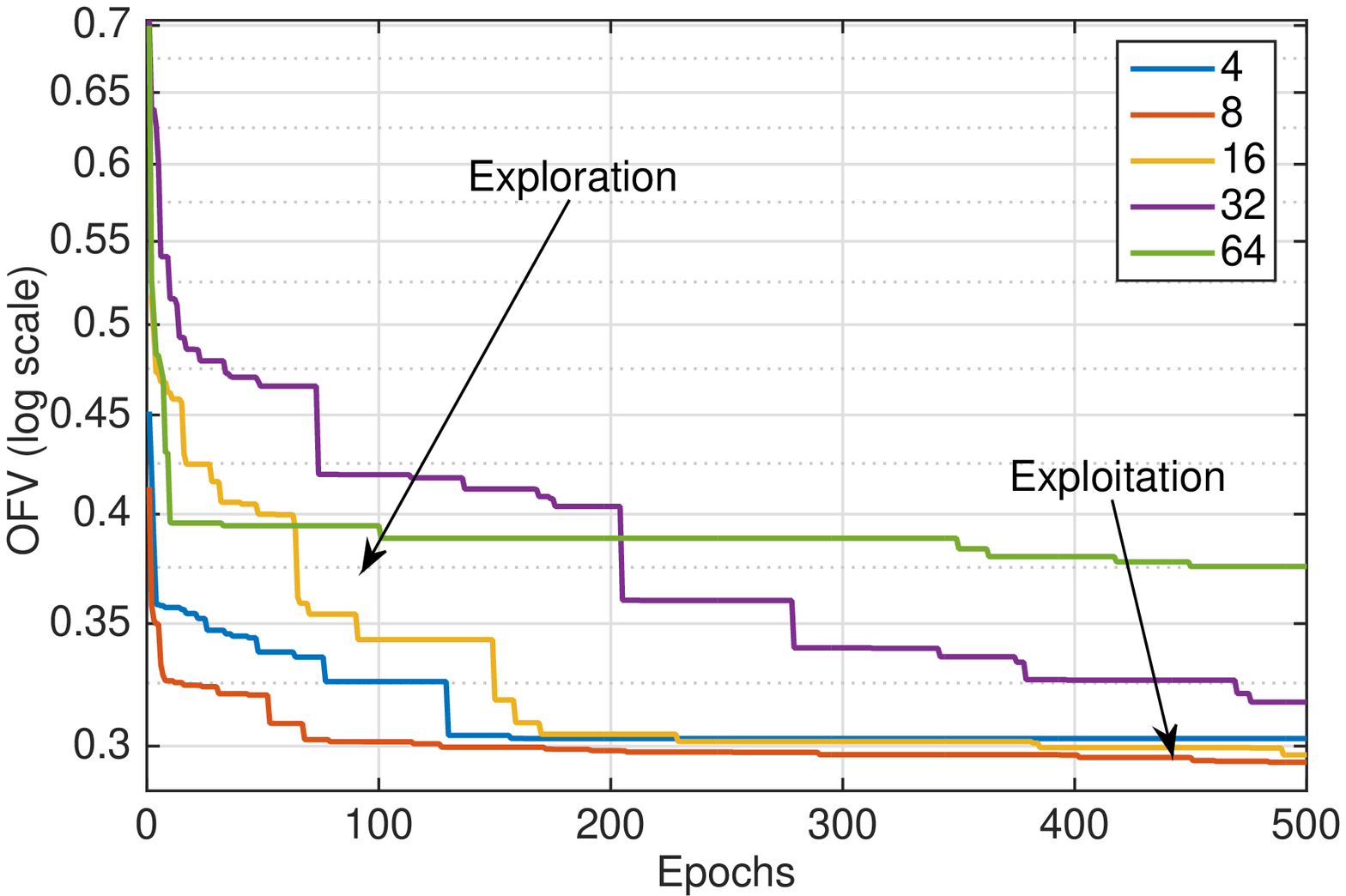}  
\caption{ArrowHead}
\label{fig:population_study_arrowhead}
\end{subfigure}%

\begin{subfigure}[t]{0.43\textwidth}
\captionsetup{font=footnotesize}
\centering
\includegraphics[width=1\textwidth]{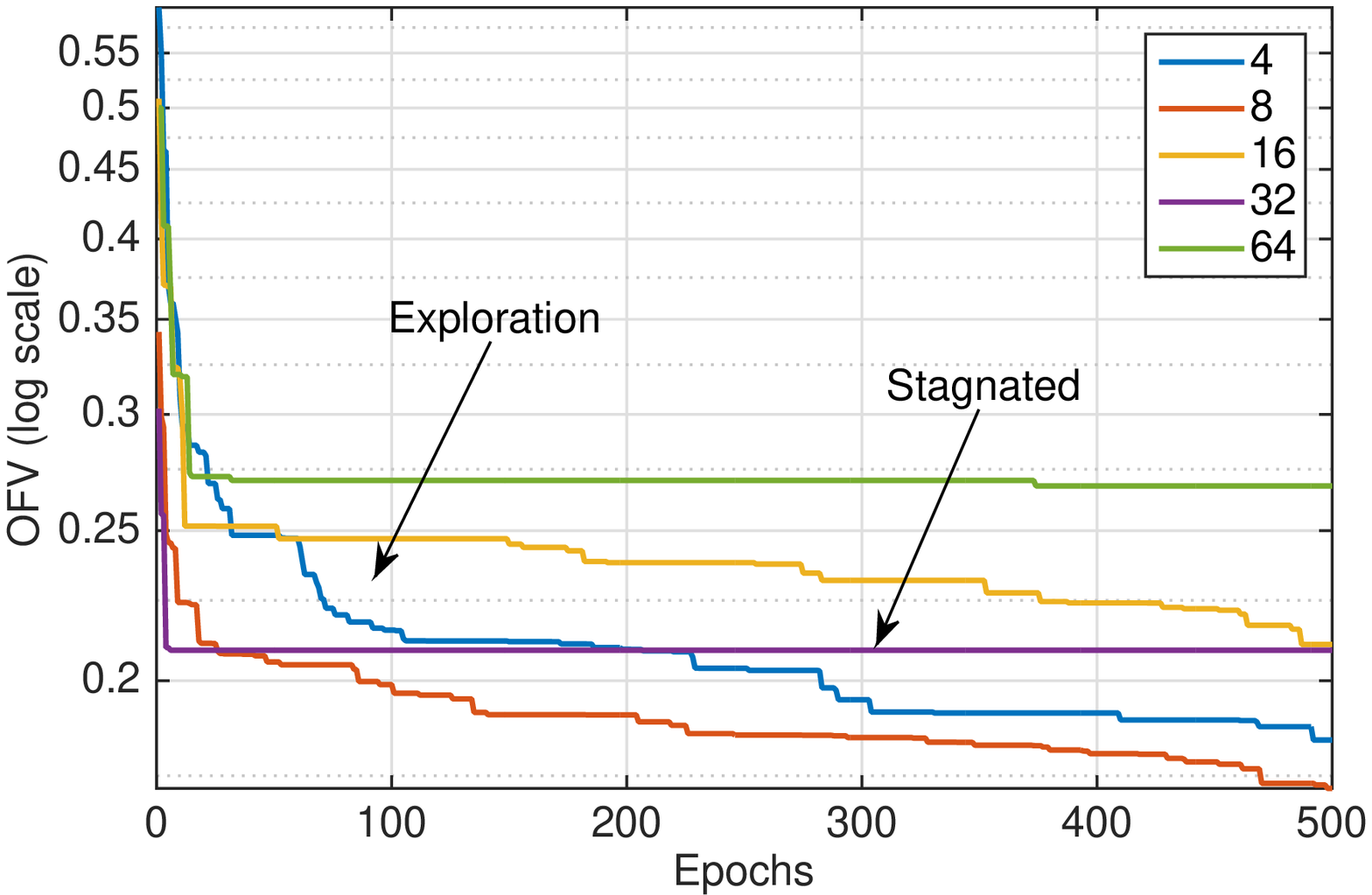}
\caption{BeetleFly}
\label{fig:population_study_beetlefly}
\end{subfigure}%
\caption{Average objective function value (OFV) of S-Rocket over $10$ independent runs for population sizes $S\in\{4,8,16,32,64\}$.} 
\label{fig:population_study}
\vspace{-4mm}
\end{figure}

\begin{figure}[!t]
\centering
\captionsetup{font=footnotesize}
\begin{subfigure}[t]{0.25\textwidth}
\captionsetup{font=footnotesize}
\centering
\includegraphics[width=1\textwidth]{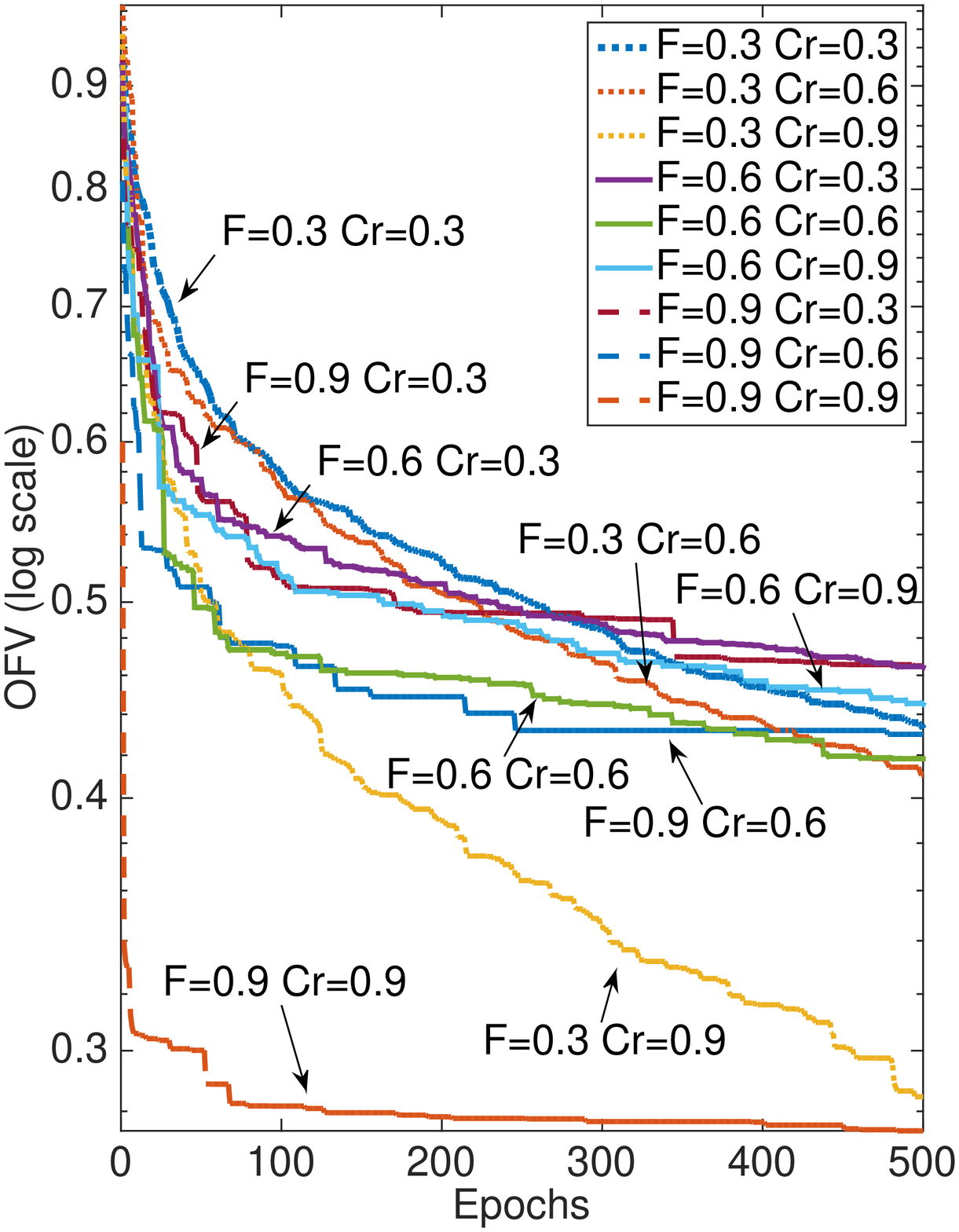}  
\caption{ArrowHead}
\label{fig:CrF_study_arrowhead}
\end{subfigure}%
\begin{subfigure}[t]{0.25\textwidth}
\captionsetup{font=footnotesize}
\centering
\includegraphics[width=1\textwidth]{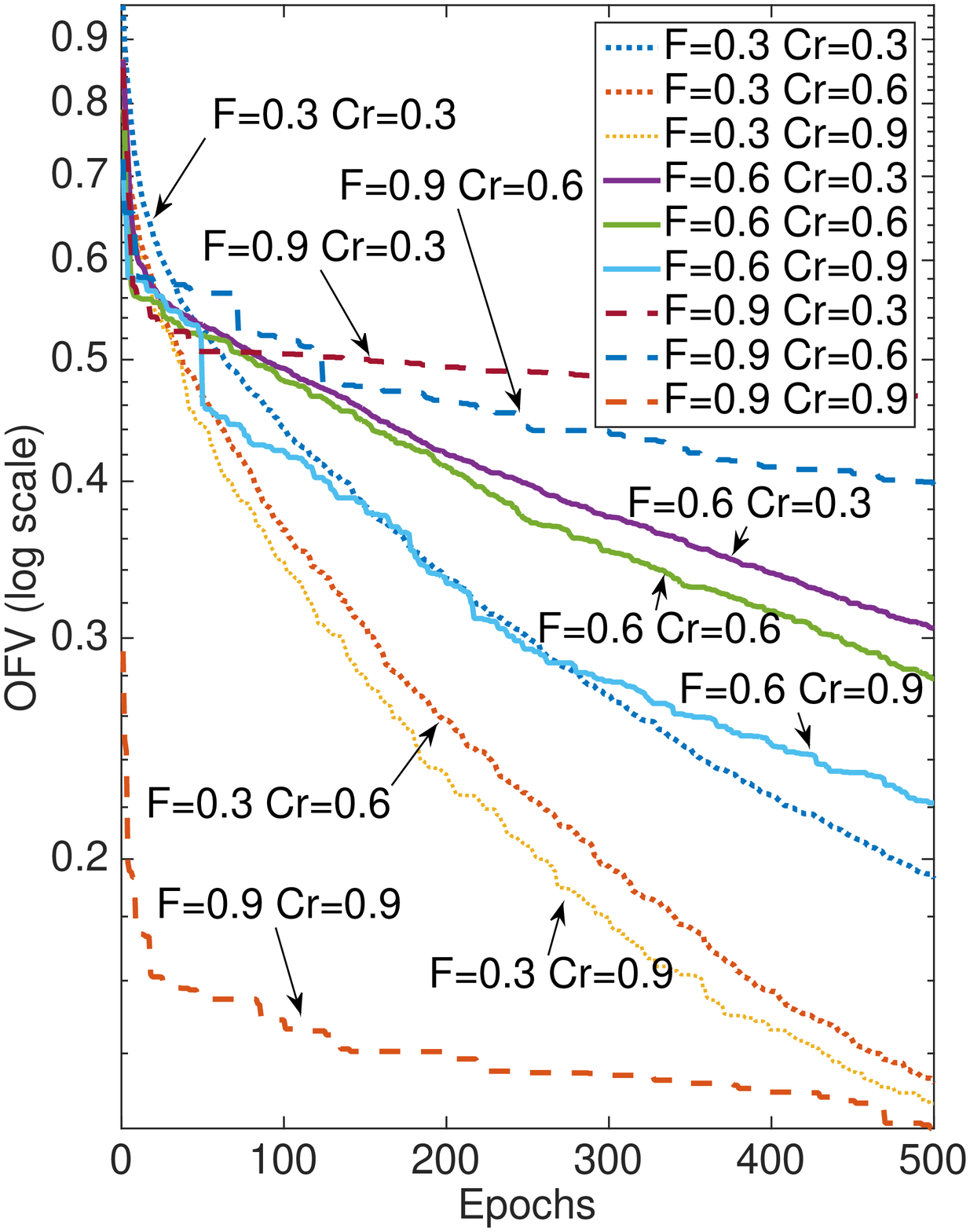}
\caption{BeetleFly}
\label{fig:CrF_study_beetlefly}
\end{subfigure}%
\caption{The objective function value (OFV) of S-Rocket for the ArrowHead and BeetleFly datasets averaged over $10$ independent runs for $F\in\{0.3,0.6,0.9\}$ and $C_r\in\{0.3,0.6,0.9\}$.} 
\label{fig:CrF_study}
\vspace{-4mm}
\end{figure}

\subsection{Post-Training}

The sparse \textit{best state vector} $\mathbf{s}_b$ represents participation status of a kernel in feature extraction, where $s_{b,d}=1$ means kernel $d$ remains active. The sparse features are used to retrain the Ridge regression classifier by solving a regression model where the loss function is the linear least squares function and regularization is given by the $l_2$-norm.


\section{Experiments}
\label{sec:experiments}

\subsection{Data}
Similar to~\cite{dempster2020rocket}, experiments are conducted on the UCR archive~\cite{dau2019ucr} time series classification datasets (the first $30$ datasets). Generally, these datasets have limited training samples with a similar/larger size test dataset. 

\subsection{Setup}
The open-source Rocket and MiniRocket codes were used for the experiments\footnote{\textit{https://github.com/angus924/rocket}}. The regularization coefficient of Ridge regression classifier was set based on a cross-validation search in the set of 10 numbers spaced evenly on the log scale range of $[-3,3]$. Our developed codes for S-Rocket are available online\footnote{\textit{https://github.com/salehinejad/srocket}}. 
In the experiments, otherwise stated, the number of random kernels is $D=10,000$~\cite{dempster2020rocket}, the number of training epochs is $500$, and the average results of $10$ independent runs are reported. The number of candidate state vectors is $S=8$ where $50\%$ of them are initialized from the Bernoulli distribution with the probability $0.5$ and the others are initialized as $1$ and then shuffled. 

\subsection{Classification Performance Analysis}
Tables~\ref{T:srocket_results} and~\ref{T:sminirocket_results} show the average classification performance (Acc.), average objective function value (OFV) according to~(\ref{eq:s_rocket_loss}), Matthews correlation coefficient (MCC), and ratio of kept kernels ($D'$) for Rocket and MiniRocket and the corresponding S-Rocket and S-MiniRocket implementation, respectively. The S-Rocket model is compared with $l_1$-norm pruning~\cite{li2016pruning}, Soft Filter pruning~\cite{he2018soft}, and a \textit{Random} mask pruning approach. In the \textit{Random} pruning approach, a random mask is applied on the kernels and the classifier is trained with the corresponding PPV values, similar to the S-Rocket setup. The pruning rate in these models is set to the $D'$ rate found by the S-Rocket model. 

From the classification accuracy perspective, S-Rocket has an equal performance in $19$, lower performance in $9$ ($\leq 3\%$), and better performance in $2$ ($\leq 1\%$) datasets. Similarly, S-MiniRocket has an equal performance in $17$, lower performance in $12$ ($\leq 4\%$), and better performance in $1$ ($\leq 1\%$) datasets. S-Rocket and S-MiniRocket on average use about $39\%$ and $36\%$ of the kernels ($D'$), respectively, to achieve an almost similar accuracy to the original model. Hence, the OFV using (\ref{eq:s_rocket_loss}) is lower for S-Rocket and S-MiniRocket in comparison with their original counterpart. Overall, the models with pruned kernels have achieved almost similar performance to the original model with less than $40\%$ of the kernels. Figure~\ref{fig:scatter_results} shows the trade-off between accuracy and the number of selected features (kernels) for all datasets in Tables~\ref{T:srocket_results} and~\ref{T:sminirocket_results}. Since the ratio of utilized kernels for Rocket and MiniRocket is $D'=1$, the corresponding indicators have a linear pattern. The center of each cluster represents the average accuracy and the average number of kept kernels over all datasets.

\subsection{Convergence Analysis}
Population-based optimization algorithms perform a global search and typically converge to a local solution. Generally, the algorithm spends a number of early epochs for exploring the optimization landscape (referred to as \textit{exploration}) and gradually moves toward fine-tuning of the solutions (referred to as \textit{exploitation}). In some cases, the population may experience \textit{premature convergence} due to lack of diversity and an early convergence to a local optima. It may also suffer from stagnation where the population remains diverse during the optimization process~\cite{lampinen2000stagnation}.

Figure~\ref{fig:convergence} shows the average of accuracy, ratio of active kernels, and objective function value using (\ref{eq:s_rocket_loss}), scaled to $[0,1]$, over $10$ independent runs and $500$ epochs for the ArrowHead and BeetleFly datasets. The plots show that, generally, the optimizer is increasing the accuracy while decreasing the number of active kernels and the objective function value. As an example, the exploration phase is observable before about epoch $80$ in Figure~\ref{fig:convergence}(a).   

 In general, a large population size enhances exploration of the optimization landscape and decreases the convergence rate. A small population size helps faster convergence but increases the risk of trapping in a local optima and pre-mature convergence~\cite{salehinejad2017micro}, \cite{salehinejad2014micro}. Figure~\ref{fig:population_study} shows the average objective function value of the S-Rocket over $10$ independent runs for population sizes $S\in\{4,8,16,32,64\}$. For the ArrowHead dataset, we can observe that the larger population sizes $S\in\{16,32,64\}$ have a slower convergence than the $S\in\{4,8\}$. Particularly, $S=64$ has a very slow convergence rate. 
 
 After about epoch $180$, the $S\in\{4,8,16\}$ are transiting from exploration to exploitation and a slow improvement in the OFV is observable. The small populations $S\in\{4,8\}$ are relatively closer values but the $S=8$ has a better balance of exploration and exploitation and can achieve a lower OFV. A similar behaviour is observable for the BeetleFly dataset.

The mutation factor $F$ in ($\ref{eq:mutation}$) and the cross-over rate $C_r$ in ($\ref{eq:crossover}$) are other parameters to control the diversity of search and convergence rate~\cite{salehinejad2017micro}. A larger mutation factor $0\leq F\leq 1$ increases the probability of flipping a state in ($\ref{eq:mutation}$). Similarly, a larger cross-over rate $0\leq C_r \leq 1$ increases the chance of using a mutated vector and smaller value increases the chance of inheritance from the previous generation. Figure~\ref{fig:CrF_study} shows the objective function values of S-Rocket for the ArrowHead and BeetleFly datasets averaged over $10$ independent runs for $S=8$, $F\in\{0.3,0.6,0.9\}$, and $C_r\in\{0.3,0.6,0.9\}$. These plots show that a combination of larger mutation factor and cross-over rates (i.e. $C_r=0.9$ and $F=0.9$) significantly accelerate convergence of the optimizer to a lower objective function value.

\begin{figure}[!t]
\centering
\captionsetup{font=footnotesize}
\begin{subfigure}[t]{0.45\textwidth}
\captionsetup{font=footnotesize}
\centering
\includegraphics[width=1\textwidth]{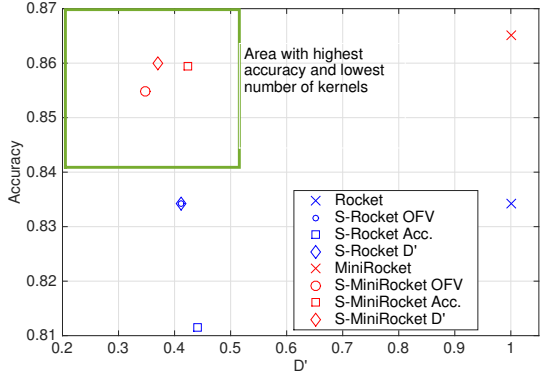} 
\caption{ArrowHead}
\label{fig:pareto_Cin}
\end{subfigure}%

\begin{subfigure}[t]{0.45\textwidth}
\captionsetup{font=footnotesize}
\centering
\includegraphics[width=1\textwidth]{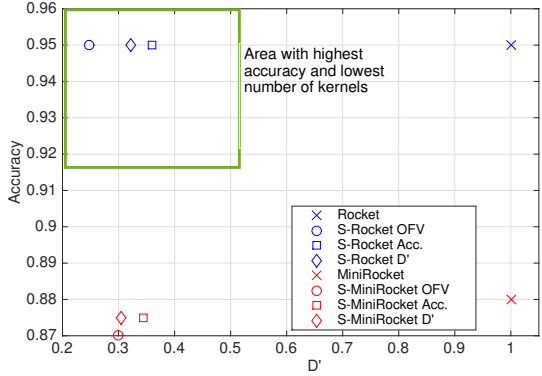}
\caption{BeetleFly}
\label{fig:pareto_computers}
\end{subfigure}%
\caption{Visualization of the classification accuracy vs. ratio of active kernels ($D'$) for the ArrowHead and BeetleFly datasets. OFV: Using the state vector with lowest objective function value according to~(\ref{eq:s_rocket_loss}); Acc.: Using the state vector with highest classification accuracy; $D'$: Using the state vector with lowest number of selected convolution kernels.
} 
\label{fig:paretofronts}
\vspace{-4mm}
\end{figure}

\begin{figure}[!t]
\centering
\captionsetup{font=footnotesize}
\begin{subfigure}[t]{0.432\textwidth}
\captionsetup{font=footnotesize}
\centering
\includegraphics[width=1\textwidth]{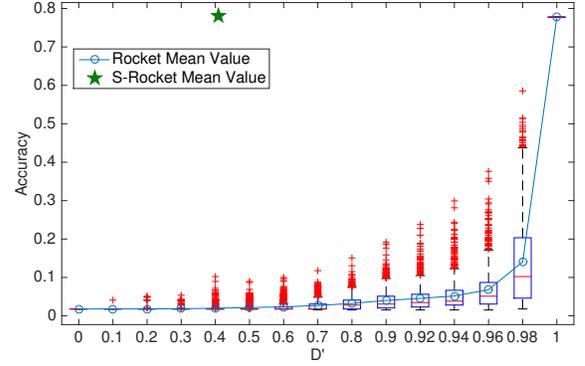} 
\caption{Adiac}
\label{fig:}
\end{subfigure}%

\begin{subfigure}[t]{0.43\textwidth}
\captionsetup{font=footnotesize}
\centering
\includegraphics[width=1\textwidth]{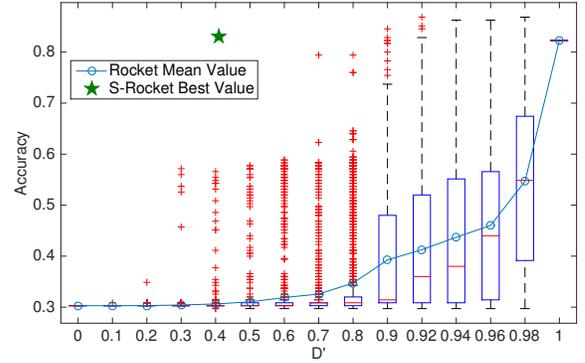}
\caption{ArrowHead}
\label{fig:}
\end{subfigure}%

\begin{subfigure}[t]{0.43\textwidth}
\captionsetup{font=footnotesize}
\centering
\includegraphics[width=1\textwidth]{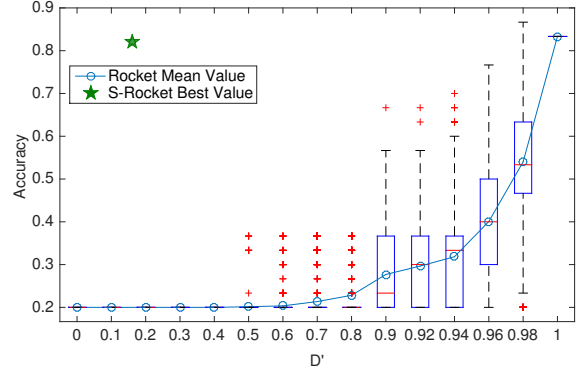}
\caption{Beef}
\label{fig:}
\end{subfigure}%

\begin{subfigure}[t]{0.43\textwidth}
\captionsetup{font=footnotesize}
\centering
\includegraphics[width=1\textwidth]{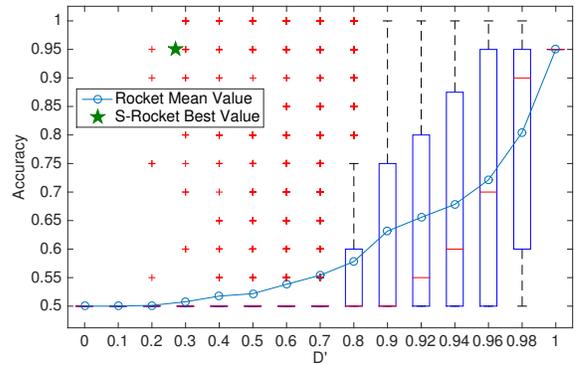}
\caption{BeetleFly}
\label{fig:}
\end{subfigure}%
\caption{Monte Carlo simulation results of classification accuracy values for different ratios of active kernels ($D'$). $1,000$ random evaluations per $D'$ have been conducted. The central mark of each box indicates the median, and the bottom and top edges indicate the $25th$ and $75th$ percentiles, respectively. The outliers are plotted individually using a red $+$ marker and the mean value is denoted by a blue $\circ$ for each $D'$.} 
\label{fig:montecarlo}
\vspace{-4mm}
\end{figure}

\begin{figure*}[!t]
\centering
\captionsetup{font=footnotesize}
\begin{subfigure}[t]{0.25\textwidth}
\captionsetup{font=footnotesize}
\centering
\includegraphics[width=1\textwidth]{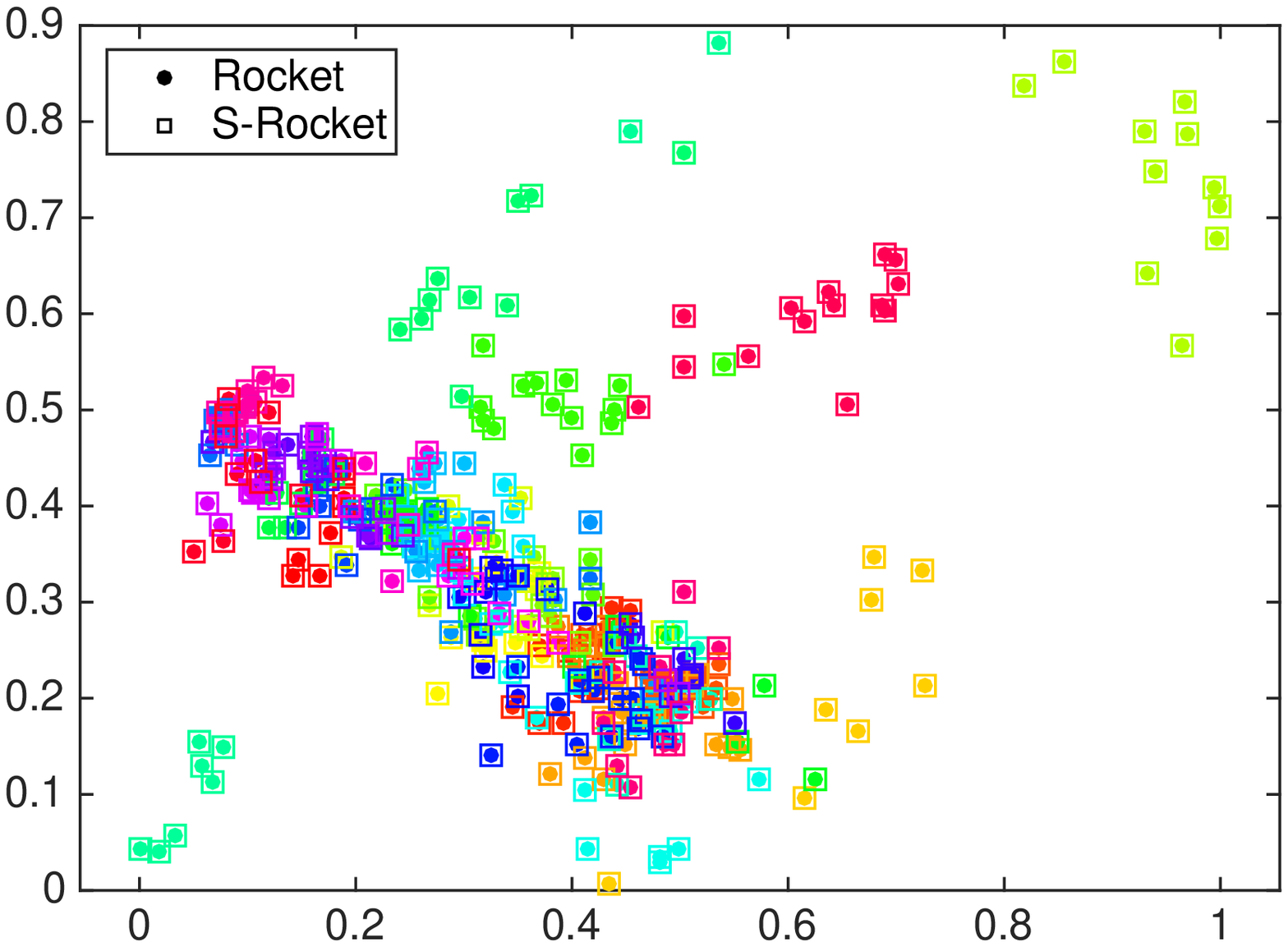}
\caption{Adiac}
\label{fig:pareto_computers}
\end{subfigure}%
\begin{subfigure}[t]{0.25\textwidth}
\captionsetup{font=footnotesize}
\centering
\includegraphics[width=1\textwidth]{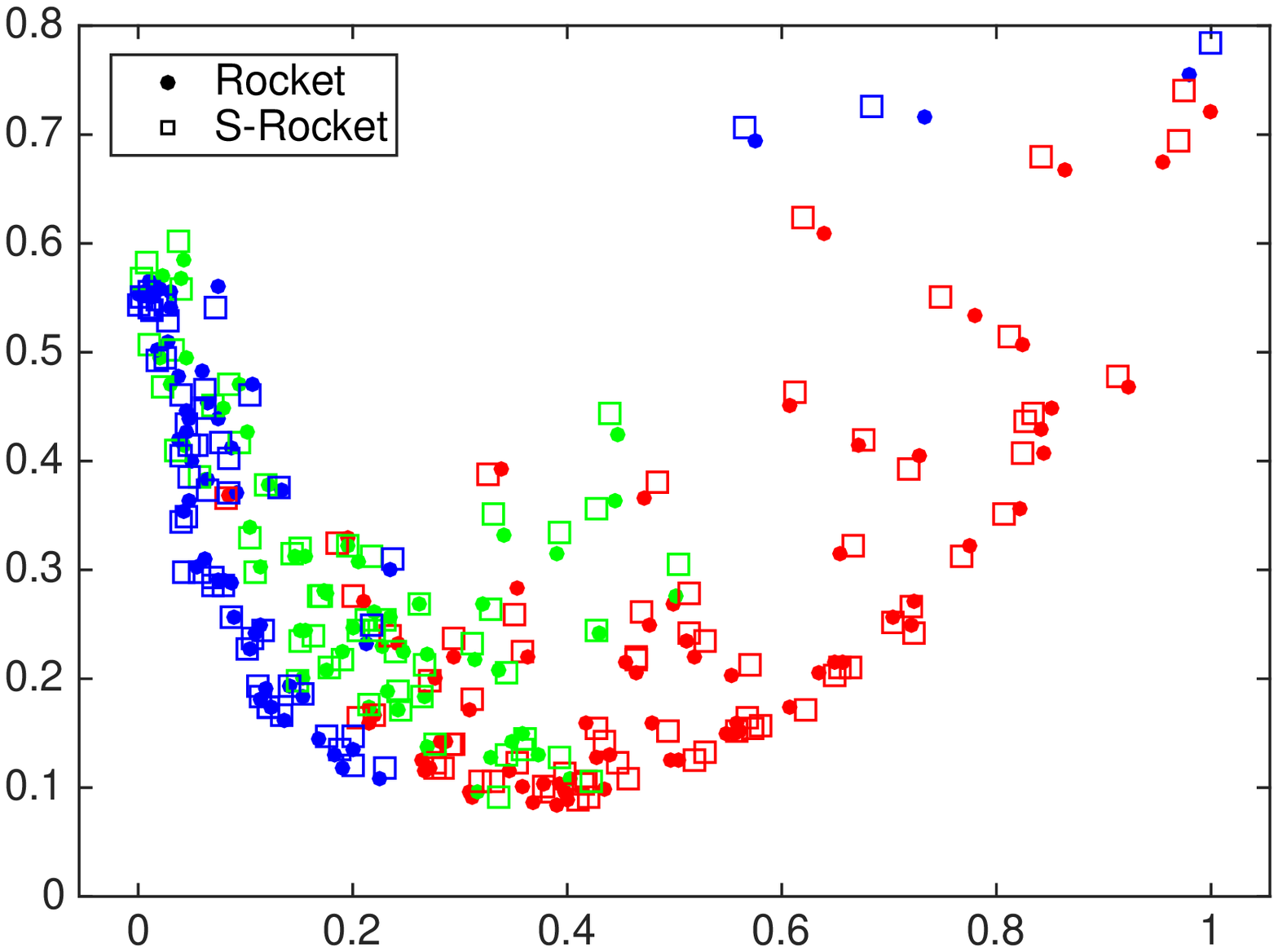} 
\caption{ArrowHead}
\label{fig:featuresrArrowHead}
\end{subfigure}%
\begin{subfigure}[t]{0.25\textwidth}
\captionsetup{font=footnotesize}
\centering
\includegraphics[width=1\textwidth]{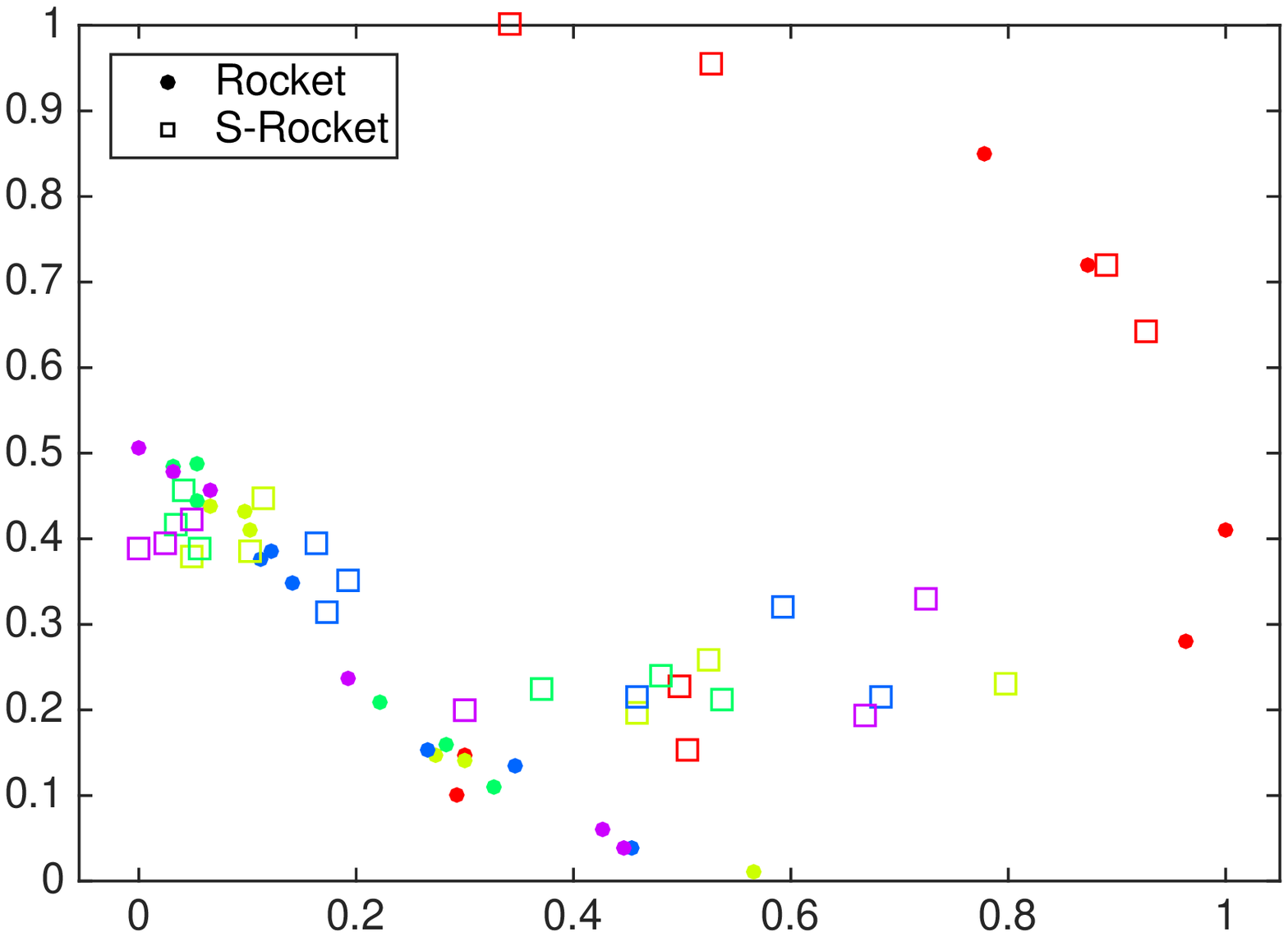}
\caption{Beef}
\end{subfigure}%
\begin{subfigure}[t]{0.25\textwidth}
\captionsetup{font=footnotesize}
\centering
\includegraphics[width=1\textwidth]{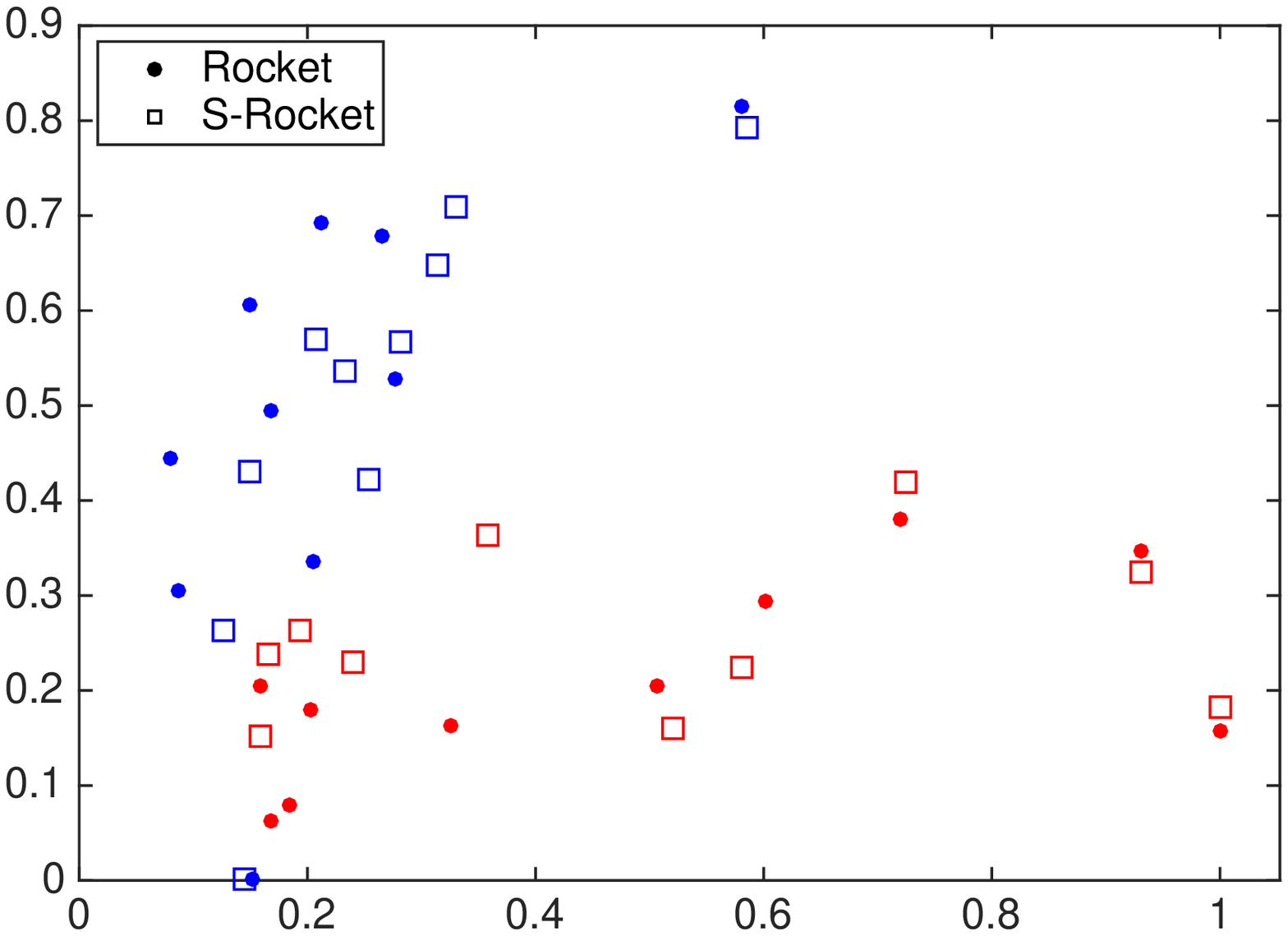}
\caption{BeetleFly}
\end{subfigure}%
\caption{Normalized two-dimensional representation of extracted features from random convolution kernels ($D=10,000$) before and after pruning, using PCA for \textbf{Rocket} and \textbf{S-Rocket}. The data points represent time series samples in each dataset where the data classes are color-coded.} 
\label{fig:featurespacer}
\end{figure*}

\begin{figure*}[!t]
\centering
\captionsetup{font=footnotesize}
\begin{subfigure}[t]{0.25\textwidth}
\captionsetup{font=footnotesize}
\centering
\includegraphics[width=1\textwidth]{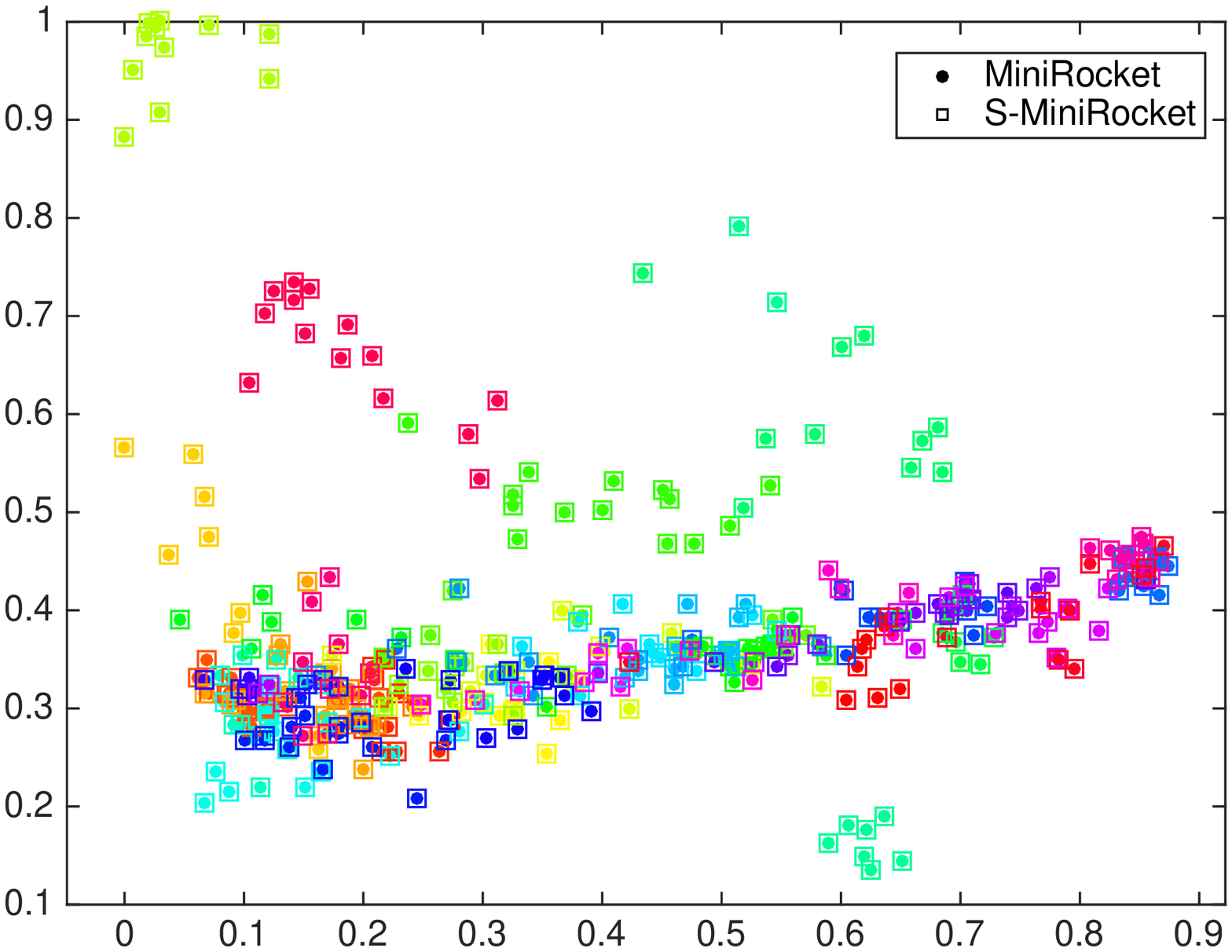}
\caption{Adiac}
\end{subfigure}%
\begin{subfigure}[t]{0.25\textwidth}
\captionsetup{font=footnotesize}
\centering
\includegraphics[width=1\textwidth]{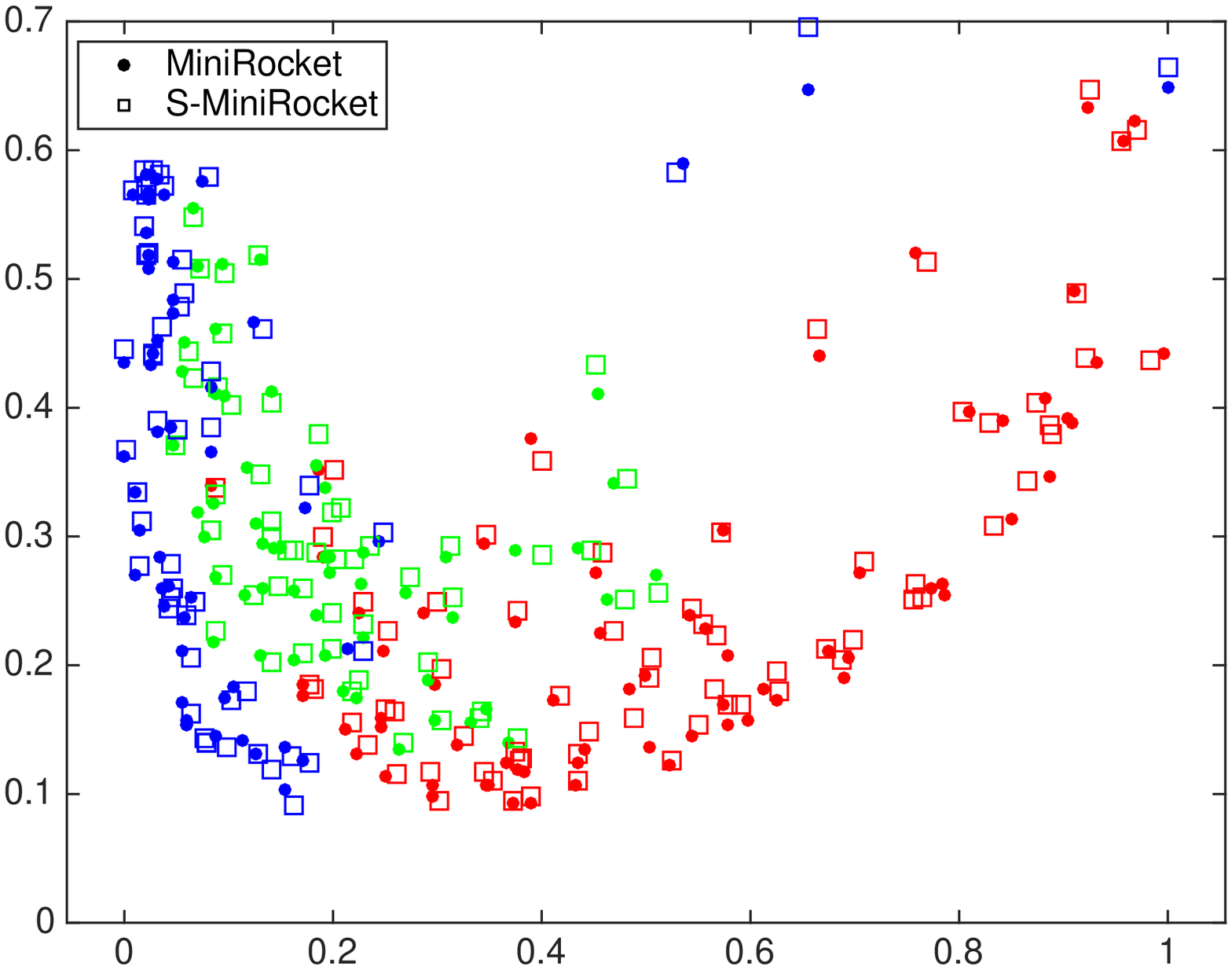}  
\caption{ArrowHead}
\end{subfigure}%
\begin{subfigure}[t]{0.25\textwidth}
\captionsetup{font=footnotesize}
\centering
\includegraphics[width=1\textwidth]{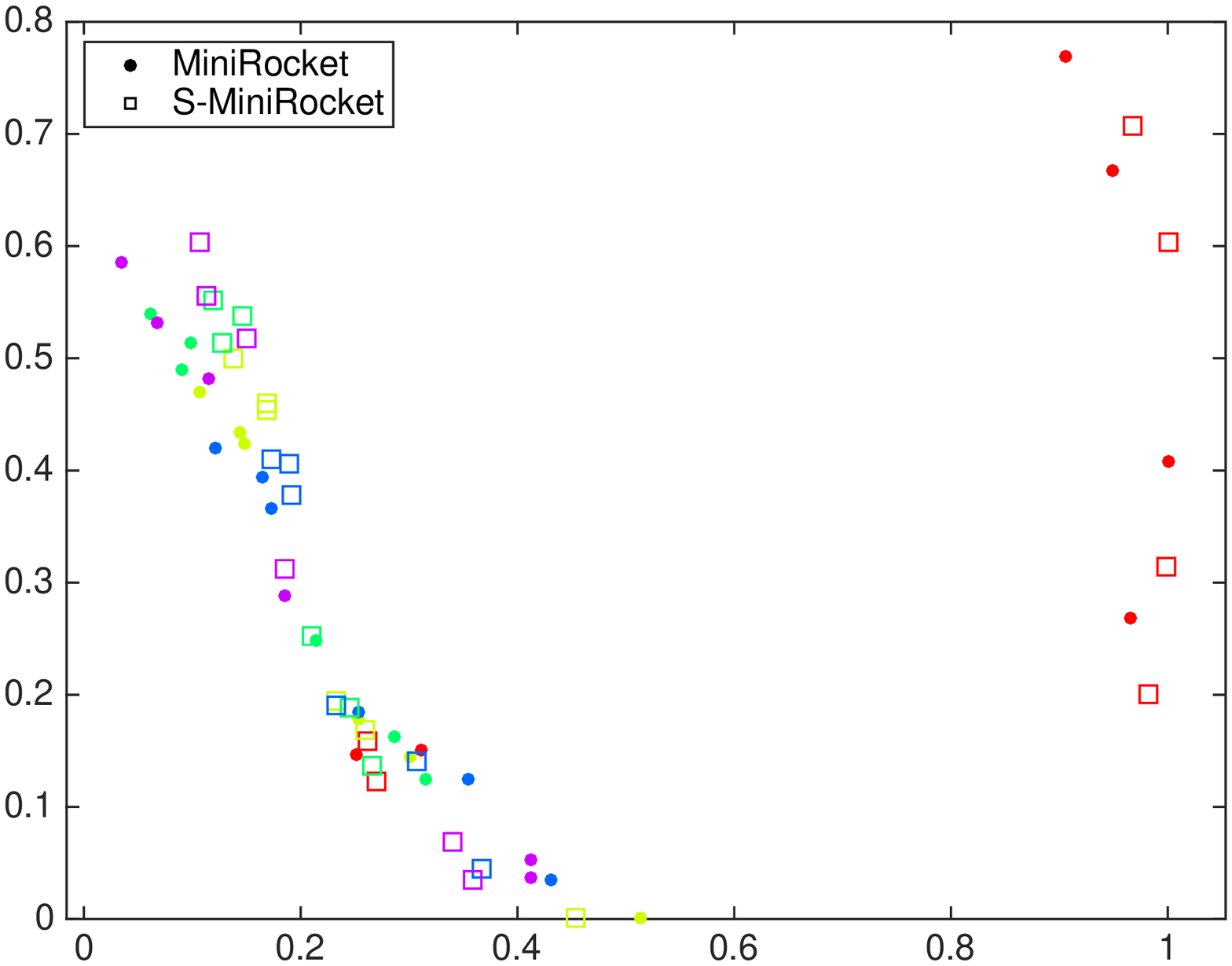}
\caption{Beef}
\end{subfigure}%
\begin{subfigure}[t]{0.25\textwidth}
\captionsetup{font=footnotesize}
\centering
\includegraphics[width=1\textwidth]{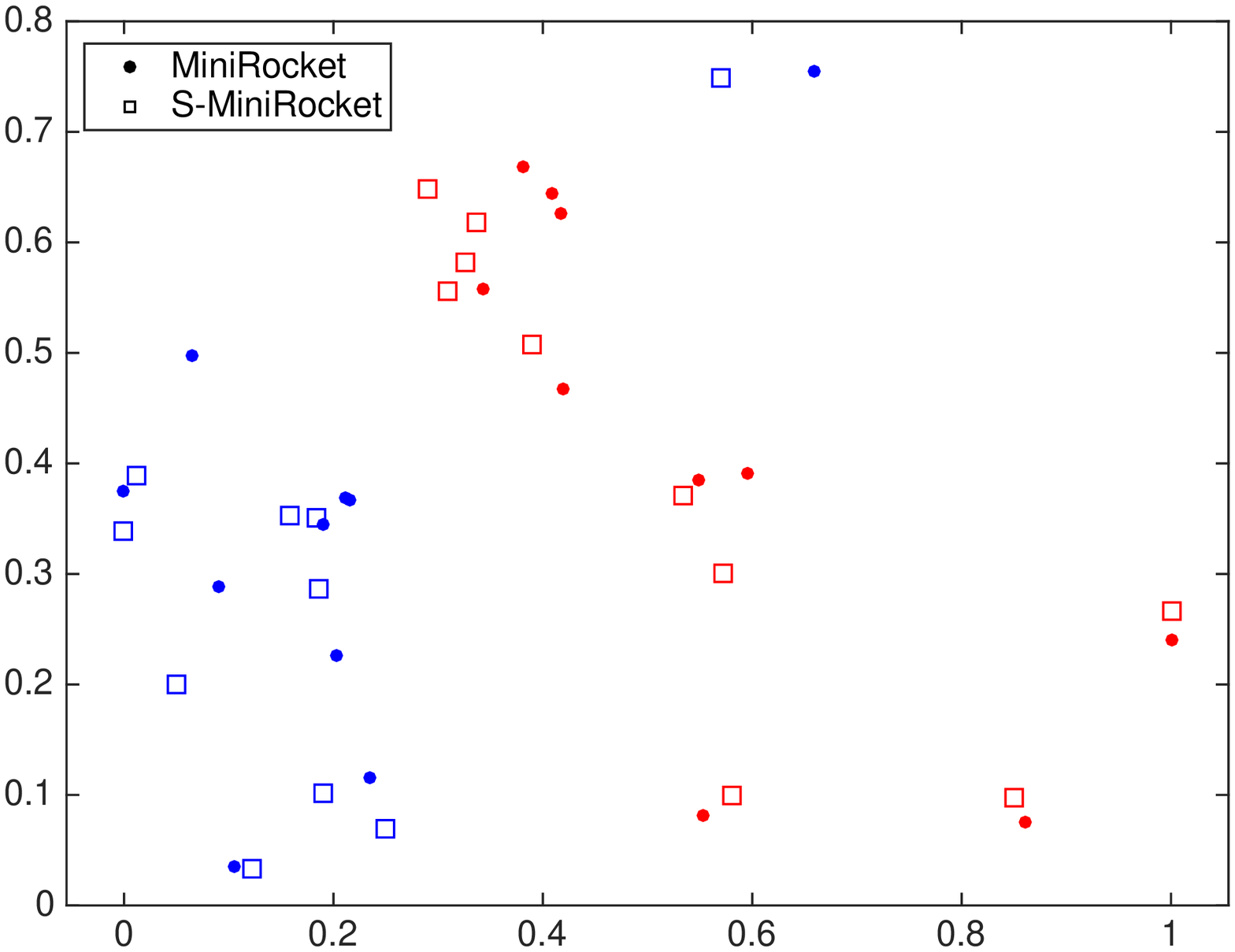}
\caption{BeetleFLy}
\end{subfigure}%
\caption{Normalized two-dimensional representation of extracted features from random convolution kernels ($D=10,000$) before and after pruning, using PCA for \textbf{MiniRocket} and \textbf{S-MiniRocket}. The data points represent time series samples in each dataset where the data classes are color-coded.}
\label{fig:featurespaceminir2}
\end{figure*}

\begin{figure}[!t]
\centering
\captionsetup{font=footnotesize}
\begin{subfigure}[t]{0.43\textwidth}
\captionsetup{font=footnotesize}
\centering
\includegraphics[width=1\textwidth]{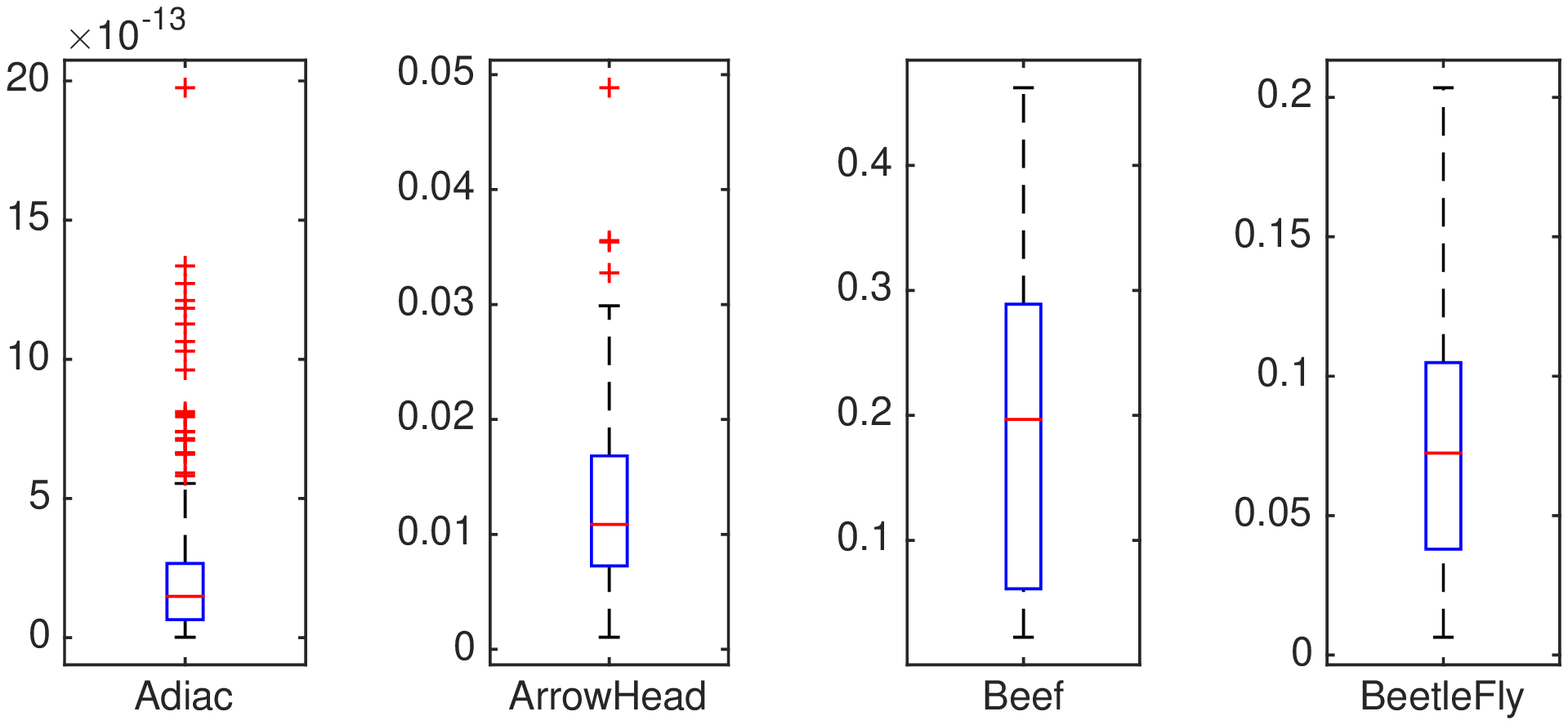}  
\caption{Rocket}
\label{fig:featurespacestatRocket}
\end{subfigure}%

\begin{subfigure}[t]{0.43\textwidth}
\captionsetup{font=footnotesize}
\centering
\includegraphics[width=1\textwidth]{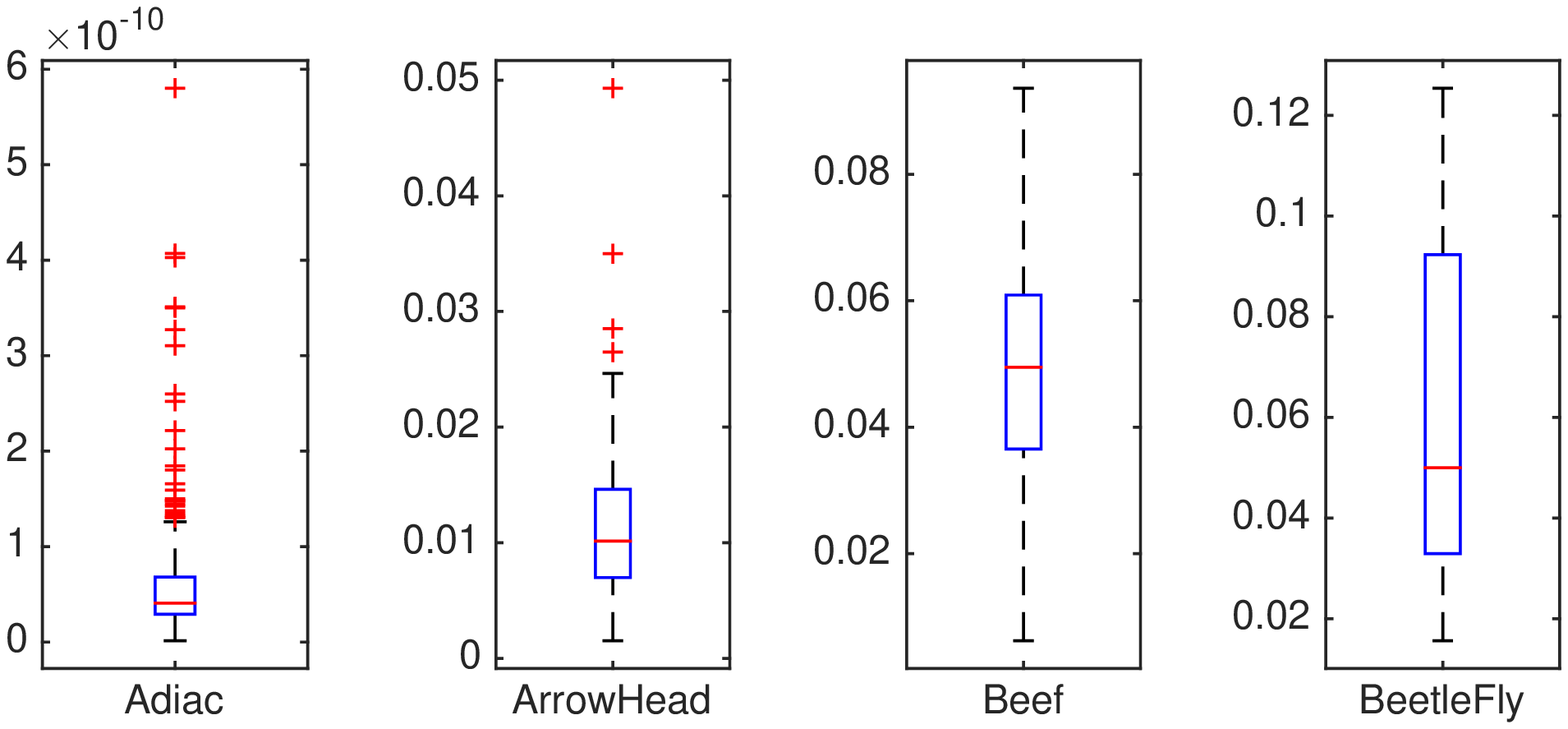}
\caption{MiniRocket}
\label{fig:featurespacestatMini}
\end{subfigure}%
\caption{Normalized Euclidean distance between two-dimensional representation of extracted features from random convolution kernels ($D=10,000$) before and after pruning, using PCA. The inputs are time series samples in each dataset.} 
\label{fig:featurespacestat}
\vspace{-4mm}
\end{figure}

\subsection{Monte Carlo Simulations}

S-Rocket reduces complexity of Rocket by pruning less important convolution kernels while maintaining the classification accuracy. Hence, there is a trade-off between these two objectives. Figure~\ref{fig:paretofronts} shows a visualization of S-Rocket and S-MiniRocket for the ArrowHead and BeetleFly datasets. A desired area is where the accuracy is maximized while the number of input features is minimized (in green). 
The \textit{S-Rocket/S-MiniRocket OFV} marker represents the best model during optimization with respect to the accuracy and the number of active kernels using~(\ref{eq:s_rocket_loss}). The \textit{S-Rocket/S-MiniRocket Acc.} marker represents the best model during optimization with respect to maximizing the accuracy. The \textit{S-Rocket/S-MiniRocket $D'$} marker represents the best model during optimization with respect to minimizing the number of kernels.  

Figure~\ref{fig:montecarlo} shows Monte Carlo simulation results of classification accuracy values for different arbitrary ratios of active kernels ($D'$). For each $D'$, $1,000$ random state vectors (without replacement) from a standard uniform distribution are generated. Then, each vector $\mathbf{s}_i$ is applied to the extracted features from random kernels of a trained Rocket model as $\mathbf{s}_{i}\odot \mathbf{k}_n$, where $\mathbf{k}_n$ is extracted from the test dataset. These figures show efficiency of the S-Rocket in finding a state vector which maximizes the accuracy while minimizing the number of active kernels. For instance in Figure~\ref{fig:montecarlo}(c), the accuracy of S-Rocket is $82\%$ and $16\%$ active kernels (green $\star$), which is equivalent to the Rocket at full capacity (i.e. $100\%$ active kernels).

\begin{table*}[!t]
\captionsetup{font=footnotesize}
\caption{Execution time of different steps of Rocket and S-Rocket in Seconds. \textit{Initialization \& Convolution} step is a common step in both models. The \textit{Training/Pre-Training} step refers to the Rocket/S-Rocket models.
}
\centering
\begin{adjustbox}{width=0.8\textwidth}
\centering

\begin{tabular}{|c|cc||c||ccc|}
\hline
\multirow{2}{*}{Dataset} &
  \multicolumn{2}{c||}{Common Steps (in Seconds)} &
  \multicolumn{1}{c||}{Rocket (in Seconds)} &
  \multicolumn{3}{c|}{S-Rocket (in Seconds)} \\ \cline{2-7} 
 &
  \multicolumn{1}{c|}{\begin{tabular}[c]{@{}c@{}}Initialization \\ \& Convolution\end{tabular}} &
  \begin{tabular}[c]{@{}c@{}}Training/\\ Pre-Training\end{tabular} &
  \multicolumn{1}{c||}{Inference} &
  
  \multicolumn{1}{c|}{\begin{tabular}[c]{@{}c@{}}Optimization\\ (per epoch)\end{tabular}} &
  \multicolumn{1}{c|}{Post-Training} &
  \multicolumn{1}{c|}{Inference} \\ \hline

Adiac &
  \multicolumn{1}{c|}{4.020} &
  5.320 &
  \multicolumn{1}{c||}{0.806} &
  \multicolumn{1}{c|}{1.067} &
  \multicolumn{1}{c|}{4.952} &
  \multicolumn{1}{c|}{\textbf{0.660}} \\ \hline
ArrowHead &
  \multicolumn{1}{c|}{1.693} &
  0.211 &
  \multicolumn{1}{c||}{0.050} &
  \multicolumn{1}{c|}{0.027} &
  \multicolumn{1}{c|}{0.113} &
  \multicolumn{1}{c|}{\textbf{0.008}}  \\ \hline
Beef &
  \multicolumn{1}{c|}{1.401} &
  0.108 &
  \multicolumn{1}{c||}{0.023} &
  \multicolumn{1}{c|}{0.029} &
  \multicolumn{1}{c|}{0.081} &
  \multicolumn{1}{c|}{\textbf{0.015}}  \\ \hline
BeetleFly &
  \multicolumn{1}{c|}{0.893} &
  0.066 &
  \multicolumn{1}{c||}{0.034} &
  \multicolumn{1}{c|}{0.013} &
  \multicolumn{1}{c|}{0.038} &
  \multicolumn{1}{c|}{\textbf{0.004}}\\ \hline
\end{tabular}
\end{adjustbox}
\label{T:timetable}
\end{table*}

\subsection{Feature Space Analysis}
Principal components analysis (PCA)~\cite{jolliffe2005principal} and t-distributed stochastic neighbor embedding (t-SNE)~\cite{van2008visualizing} are well-established methods for studying and visualization of features in a lower dimensional space such as in human activity recognition~\cite{zhang2021widar3} and medical images analysis~\cite{salehinejad2018synthesizing}.

Figures~\ref{fig:featurespacer} and~\ref{fig:featurespaceminir2} show the extracted features from random convolution kernels after reduction and normalization to a two-dimensional space using PCA for the Rocket, S-Rocket, MiniRocket, and S-MiniRocket models using Adiac, ArrowHead, Beef, and BeetleFly datasets. These scatter plots represent the two-dimensional features of samples in each dataset, where the corresponding time series class is color-coded. For example in Figure~\ref{fig:featurespacer}(\subref{fig:featuresrArrowHead}), the ArrowHead dataset has three classes which are denoted by green, red, and blue. Figures~\ref{fig:featurespacer} shows the density of features in a lower dimensional space for Rocket and S-Rocket. The shorter the distance between S-Rocket and Rocket feature clusters per class (color coded) show closer feature extraction performance of S-Rocket to Rocker. 

In other words, it shows that S-Rocket can remote redundant and less important kernels (and hence the corresponding features), without affecting the important ones. This also supports the very close classification performance results of S-Rocket to Rocket in Table~\ref{T:srocket_results}. A similar pattern is observable in Figure~\ref{fig:featurespaceminir2} for the features extracted using MiniRocket and S-MiniRocket.

The normalized Euclidean distance between extracted features using Rocket and S-Rocket from the time series samples studied in Figure~\ref{fig:featurespacer} is plotted in Figure~\ref{fig:featurespacestat}(\subref{fig:featurespacestatRocket}). These plots show that the features after pruning less important kernels are close to the features before pruning. Figure~\ref{fig:featurespacestat}(\subref{fig:featurespacestatMini}) also shows a similar pattern for MiniRocket and S-MiniRocket.

\subsection{Complexity Analysis} 

\subsubsection{Training}
The Ridge regression classifier has a complexity of $O_{R}=O(N^{2}\cdot D)$ when $N<D$~\cite{dempster2020rocket}, \cite{dongarra2018singular}. 
The implementation of the transforms in Rocket has a computational complexity of $O_{T}=O(D\cdot N \cdot l_{input})$, where $l_{input}$ is the length of the time series~\cite{dempster2020rocket}. Hence, the complexity of Rocket is a linear function of the number of features (kernels) and its total complexity is 
\begin{equation}
O_{Rocket}^{(training)}=O(D\cdot N \cdot l_{input})+O(N^{2}\cdot D).
\end{equation}
Since S-Rocket has three steps in the training phase, its complexity is \begin{equation}
    O_{S-Rocket}=O_{Pre}+O_{Opt}+O_{Post},
\end{equation}
where 
$O_{Pre}=O_{Rocket}$ is for the pre-training and 
$O_{Opt}=O(S\cdot N^{3} \cdot D \cdot N_{epochs})$~\cite{zhong2021hybrid} is for the optimization, with the worst-case assumption of $D'=D$. The complexity of the post-training step is 
\begin{equation}
    O_{Post}=O(D'\cdot N \cdot l_{input})+O(N^{2}\cdot D'),
\end{equation}
where $D'\leq D$.
Therefore the total training cost for S-Rocket is
\begin{equation}
\begin{split}
O^{(training)}_{S-Rocket} &= 
O_{Rocket}^{(training)} + O(S\cdot N^{3} \cdot D \cdot N_{epochs})\\
 &+ \; O(D'\cdot N \cdot l_{input})+O(N^{2}\cdot D'), 
\end{split}
\end{equation}
which can be simplified to 
\begin{equation}
  O^{(training)}_{S-Rocket} \; = \;
O_{Rocket}^{(training)} + O(S\cdot N^{3} \cdot D \cdot N_{epochs}).  
\end{equation}

\subsubsection{Inference} 
In inference, the complexity of Rocket and S-Rocket for a single time series is 
\begin{equation}
    O_{Rocket}^{(inference)}=O(D\cdot l_{input})+O(D),
    \end{equation}
and 
\begin{equation}
O_{S-Rocket}^{(inference)}=O(D'\cdot l_{input})+O(D'),
\end{equation}
respectively. Since $D'\leq D$, then
$O_{S-Rocket}^{(inference)}\leq O_{Rocket}^{(inference)}$.

Table~\ref{T:timetable} presents the execution time of different steps of the Rocket and S-Rocket, which shows the inference time of S-Rocket is less than Rocket in all experiments.

\section{Conclusions}
\label{sec:conclusion}
Training a linear classifier using feature generated from a bank of random convolution kernels (without training the kernels) is a fast and efficient approach for time series classification. Rocket and MiniRocket are two methods based on this idea for time series classification. In this paper, we propose a method for pruning less efficient and redundant kernels in Rocket and MiniRocket while maintaining the classification accuracy of the original models, called S-Rocket and S-MiniRocket, respectively. This approach can reduce computational complexity of Rocket and MiniRocket in inference mode for implementation on devices with limited resources such as edge devices. The results show that S-Rocket and S-MiniRocket can prune up to $99\%$ of the random convolution kernels in some standard datasets without noticeable reduction of the classification performance. Our analysis in the feature space shows that the extracted features before and after pruning of the convolution kernels are very similar, which supports the efficiency of the proposed approach in removing unnecessary kernels without affecting the performance of classifier.

\section{Acknowledgment}
This work was partially supported by the Mobile AI Lab established
between Huawei Technologies Co. LTD Canada and The Governing Council of the
University of Toronto.

\bibliographystyle{IEEEtran}
\bibliography{CTLIEEEtrans}
\end{document}